\documentclass[10pt,twocolumn,letterpaper]{article}

\usepackage{iccv}
\usepackage{times}
\usepackage{epsfig}
\usepackage{graphicx}
\usepackage{amsmath}
\usepackage{amssymb}
\usepackage{xcolor}
\usepackage{booktabs} %

\usepackage{afterpage}

\usepackage{balance}

\usepackage{amssymb}

\usepackage{setspace}

\usepackage[switch]{lineno}

\usepackage{multirow}
\usepackage{rotating}
\usepackage{float}
\usepackage[]{subfig}%
\definecolor{Gray}{gray}{0.8}
\usepackage{colortbl}
\usepackage{hhline}
\usepackage{pifont} %
\usepackage[final]{pdfpages}

\newcommand{\cmark}{\color{green}\ding{51}}
\newcommand{\xmark}{\color{red}\ding{55}}

\newcommand{\smplHF}{SMPL-X\xspace}

\newcommand{\smplifyPP}{SMPLify-X\xspace}

\newcommand{\caesar}{CAESAR\xspace}

\newcommand{\openpose}{OpenPose\xspace}

\newcommand{\pytorch}{PyTorch\xspace}
\newcommand{\mocap}{MoCap\xspace}

\newcommand{\twoD}{2D\xspace}
\newcommand{\threeD}{3D\xspace}
\newcommand{\gt}{ground-truth\xspace}

\newcommand{\mExpr}{\mathcal{E}}
\newcommand{\mColl}{\mathcal{P}} %
\newcommand{\contact}{\mathcal{C}}

\newcommand{\Jest}{J_{est}}

\newcommand{\para}{\theta}

\newcommand{\skanect}{Skanect\xspace}
\newcommand{\structureIO}{Structure Sensor\xspace}
\newcommand{\kinectTWO}{Kinect-One\xspace}
\newcommand{\pigraph}{PiGraphs\xspace}

\newcommand{\vicon}{Vicon\xspace}

\newcommand{\moshPP}{MoSh++\xspace}

\newcommand{\rgb}{RGB\xspace}
\newcommand{\rgbD}{\mbox{RGB-D}\xspace}

\newcommand{\mesh}{M}
\newcommand{\scene}{s}
\newcommand{\cam}{c}
\newcommand{\body}{b}
\newcommand{\face}{f}
\newcommand{\hand}{h}
\newcommand{\imgRGB}{I}
\newcommand{\imgDDD}{Z}

\newcommand{\trans}{{\gamma}}

\newcommand{\annealingGamma}{\kappa}
\newcommand{\colisSelfTR}{{self}}
\newcommand{\colisInterTR}{{inter}}

\newcommand{\ncomps}{{12}}
\newcommand{\ourDataSUBJECTS}{$20$\xspace}
\newcommand{\ourDataSCENES}{$12$\xspace}
\newcommand{\citeMOSH}{\cite{AMASS:2019}\xspace}

\newcommand{\proxD}{{\mbox{PROX-D}}\xspace}
\newcommand{\SD}{{\mbox{SMPLify-D}}\xspace}
\newcommand{\prox}{\mbox{PROX}\xspace}

\newcommand{\meshScene}{{M_\scene}\xspace}
\newcommand{\ourViconFrames}{$180$\xspace}

\newcommand{\supmat}{{{Sup.~Mat.}\xspace}}

\DeclareSymbolFont{matha}{OML}{txmi}{m}{it}%
\DeclareMathSymbol{\viz}{\mathord}{matha}{118}
\newcommand{\visible}{\viz}

\usepackage[pagebackref=true,breaklinks=true,letterpaper=true,colorlinks,bookmarks=false]{hyperref}

\iccvfinalcopy %

\def\iccvPaperID{2115} %

\ificcvfinal\fi

\begin{document}
\title{Resolving 3D Human Pose Ambiguities with 3D Scene Constraints}

\author{Mohamed Hassan, Vasileios Choutas, Dimitrios Tzionas and Michael J. Black\\
	Max Planck Institute for Intelligent Systems\\
	{\tt\small \{mhassan, vchoutas, dtzionas, black\}@tuebingen.mpg.de}
}

\ificcvfinal\thispagestyle{empty}\fi

\twocolumn[
{%

    \renewcommand\twocolumn[1][]{#1}%
    \maketitle
    \centering
    	\ificcvfinal\thispagestyle{empty}\fi
    \begin{minipage}{0.98\textwidth}
        \includegraphics[trim=000mm 000mm 000mm 000mm, clip=false, width=1.00\textwidth, keepaspectratio]{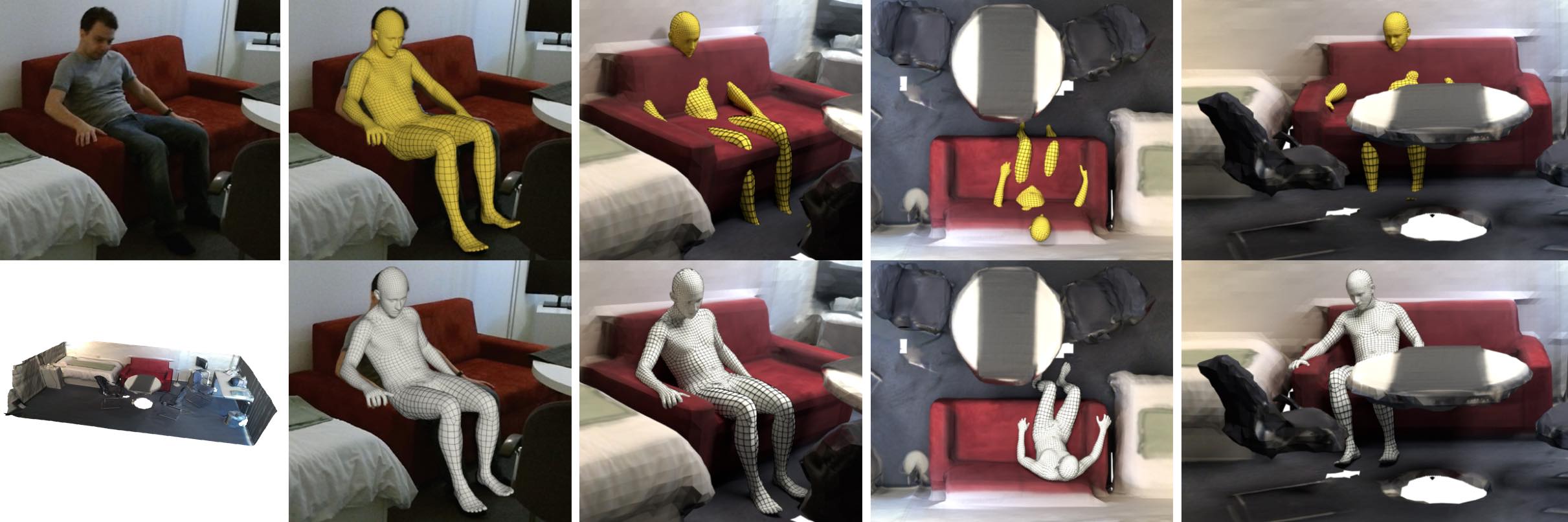}
    \end{minipage}
    \captionof{figure}{
        Standard \threeD body estimation methods predict bodies that
        may be inconsistent with the \threeD scene even though the
        results may look reasonable from the camera viewpoint.
        To address this, we exploit the \threeD scene structure and introduce \emph{scene constraints} for \emph{contact} and \emph{inter-penetration}. From left to right:
        (1) \rgb image (top) and \threeD scene reconstruction (bottom),
        (2) overlay of estimated bodies on the original \rgb image without (yellow) and with (gray) scene constraints,
        \threeD rendering of both the body and the scene from
        (3) camera view,
        (4) top view and
        (5) side view.
    }\label{fig:teaser}
    \vspace*{+08.00mm}
}
]%

\begin{abstract}

To understand and analyze human behavior, we need to capture humans moving in, and interacting with, the world. 
Most existing methods perform \threeD human pose estimation without explicitly considering the scene. 
We observe however that the world constrains the body and vice-versa. 
To motivate this, we show that current \threeD human pose estimation methods produce results that are not consistent with the \threeD scene. 
Our key contribution is to exploit static \threeD scene structure to better estimate human pose from monocular images. 
The method enforces {\em Proximal Relationships with Object eXclusion} and is called {\em \prox}.
To test this, we collect a new dataset composed of  %
$12$ different \threeD scenes and  \rgb sequences of $20$ subjects moving in and interacting with the scenes. 
We represent human pose using the \threeD human body model \smplHF and extend \smplifyPP to estimate body pose using scene constraints. 
We make use of the \threeD scene information by formulating two main constraints. 
The inter-penetration constraint penalizes intersection between the body model and the surrounding \threeD scene. 
The contact constraint encourages specific parts of the body to be in contact with scene surfaces if they are close enough in distance and orientation. 
For quantitative evaluation we capture a separate dataset with \ourViconFrames \rgb frames in which the ground-truth body pose is estimated using a motion capture system.
We show quantitatively that introducing scene constraints significantly reduces \threeD joint error and vertex error. 
Our code and data are available for research at \url{https://prox.is.tue.mpg.de}.
\end{abstract}

\vspace*{-08.00mm}

\section{Introduction}

Humans move through, and interact with, the \threeD world. 
The world limits this movement and provides opportunities (affordances) \cite{gibson1950perception}.
In fact, it is through contact between our feet and the environment that we are able to move at all.
Whether simply standing, sitting, lying down, walking, or manipulating objects, our posture, movement, and behavior is affected by the world around us.
Despite this, most work on \threeD human pose estimation from images ignores the world and our interactions with it.

Here we formulate human pose estimation differently, making the \threeD world a first class player in the solution.
Specifically we estimate \threeD human pose from a {\em single RGB image} conditioned on the \threeD scene.  
We show that the world provides constraints that make the \threeD pose estimation problem easier and the results more accurate.

We follow two key principles to estimate \threeD pose in the context of a \threeD scene.
First, from intuitive physics, two objects in \threeD space cannot \emph{inter-penetrate} and share the same space.
Thus, we penalize poses in which the body inter-penetrates scene objects.
We formulate this ``exclusion principle'' as a differentiable loss function that we incorporate into the \smplifyPP  pose estimation method \cite{smplifyPP}.

Second,  physical interaction requires \emph{contact} in \threeD space to apply forces.
To exploit this, we use the simple heuristic that certain areas of the body surface are the most likely to contact the scene, and that, when such body surfaces are close to scene surfaces, and have the same orientation,  they are likely to be in contact.
Although these ideas have been explored to some extent by the \threeD hand-object estimation community \cite{kyriazis2014,Oikonomidis_1hand_object,pham2018pami,Rogez:ICCV:2015,Tsoli:2018:ECCV,Tzionas:IJCV:2016} 
they have received less attention in work on \threeD body pose.
We formulate a term that implements this contact heuristic and find that it improves pose estimation.

Our method extends \smplifyPP \cite{smplifyPP}, which fits a 3D body model ``top down'' to ``bottom up'' features (e.g.~\twoD joint detections).
We choose this optimization-based framework over a direct regression method (deep neural network) because it is more straightforward to incorporate our physically-motivated constraints.
The method enforces {\em Proximal Relationships with Object eXclusion} and is called {\em \prox}.
Figure~\ref{fig:teaser} shows a representative example where
the human body pose is estimated with and without our environmental terms.
From the viewpoint of the camera, both solutions look good and match the \twoD image but, when placed in a scan of the \threeD scene, the results without environmental constraints can be grossly inaccurate.
Adding our constraints to the optimization reduces inter-penetration and encourages appropriate contact.

One may ask why such constraints are not typically used?
One key reason is that to estimate and reason about contact and inter-penetration, one needs both a model of the \threeD scene and a {\em realistic} model of the human body.
The former is easy to obtain today with many scanning technologies but, if the body model is not accurate, it does not make sense to reason about contact and inter-penetration.
Consequently we use the \smplHF body model \cite{smplifyPP}, which is realistic enough to serve as a ``proxy'' for the real human in the \threeD scene.
In particular, the feet, hands, and body of the model have realistic shape and degrees of freedom.

Here we assume that a rough \threeD model of the scene is available.
It is fair to ask whether it is realistic to perform monocular human pose estimation but assume a \threeD scene?
We argue that it is for two key reasons.
First, scanning a scene today is quite easy with commodity sensors.
If the scene is static, then it can be scanned once, enabling accurate body pose estimation from a single \rgb camera; this may be useful for surveillance, industrial, or special-effects applications.
Second, methods to estimate \threeD scene structure from a single image are advancing extremely quickly.
There are now good methods to infer \threeD depth maps from a single image \cite{Eigen:2014}, as well as methods that do more semantic analysis and estimate \threeD CAD models of the objects in the scene \cite{naseer2018indoorSurvey}.
Our work is complementary to this direction and we believe that monocular \threeD scene estimation and monocular \threeD human pose estimation should happen together.
The work here provides a clear example of why this is valuable.

To evaluate \prox, we use three datasets: two \emph{qualitative datasets} and a \emph{quantitative dataset}. 
The qualitative datasets contain: \threeD scene scans, monocular \rgbD videos and pseudo \gt human bodies. 
The pseudo \gt is extracted from \rgbD by extending \smplifyPP to %
 use both RGB and depth data to fit \smplHF.

In order to get true \gt for the quantitative dataset, we set up a living room in a marker-based motion capture environment, scan the scene, and collect \rgbD images in addition to the \mocap data.
We fit the \smplHF model to the \mocap marker data using \moshPP \citeMOSH and this provides \gt \threeD body shape and pose.
This allows us to quantitatively evaluate our method.

Our datasets and code are available for research at \url{https://prox.is.tue.mpg.de}.

\section{Related Work}			\label{sec:related}

Human pose estimation and \threeD scene reconstruction have been thoroughly studied for decades, albeit mostly disjointly.  %
Traditionally, human pose estimation methods \cite{Review_Moeslund_2006} estimate bodies in isolation ignoring the surrounding world,
while \threeD reconstruction methods focus on acquiring the dense \threeD shape of the scene only \cite{zollhofer2018sotaRconstructionRGBD} or performing semantic analysis \cite{armeni20163d,dai2017scannet,qi2017pointnet}, assuming no humans are present.
In this work we focus on exploiting and capturing human-world interactions.

The community has made significant progress on estimating human body pose and shape from images \cite{Review_Gavrila,Review_Moeslund_2006,Review_PoppeMotionAnalysis,Sarafianos:Survey:2016}. %
Recent methods based on deep learning, extend 3D human pose estimation to complex scenes \cite{kanazawa2017end,vnect,NBF:3DV:2018,Pavlakos18} but the 3D accuracy is limited.
To estimate human-scene interaction, however, more realistic body models are needed that include fully articulated hands such as in \cite{joo2018total,smplifyPP}.

{\bf Joint Human \& World Models:}
Several works focus on improving \twoD object detection, \twoD pose, and action recognition by observing \rgb imagery of people interacting with objects \cite{aksoy2010categorizing,gupta2009observing,koppula2013learning,pirsiavash2012detecting,yao2010modeling}.
\cite{delaitre2012scene,fouhey2014people,gupta2011workspace} use similar observations to reason about the \threeD scene, \ie rough \threeD reconstruction and affordances,
however scene cues are not used as feedback to improve human pose.
Another direction models human-scene interactions by hallucinating synthetic people either
in real \rgb images of scenes \cite{jiang2013hallucinated} for general scene labeling, or
in synthetic \threeD scenes to learn affordances \cite{grabner2011chair,shape2pose2014} or \threeD object layout in the scene \cite{Jiang:2012:objectArrangements}, or
in real \threeD scans of scenes \cite{fisher2015activity} for scene synthesis.
Here we exploit this \threeD structure to better capture poses of humans in it. %
In the following we focus on the more recent works of \cite{grabner2011chair,shape2pose2014,iMapper2018,scenegrok2014savva,savva2016pigraphs} that follow this idea. %

Several of these observe real human-world interactions in \rgbD videos \cite{iMapper2018,scenegrok2014savva,savva2016pigraphs}.
\cite{savva2016pigraphs} learns a joint probabilistic model over \threeD human poses and \threeD object arrangements, encoded as a set of human-centric prototypical interaction graphs (\pigraph).
The learned \pigraph can then be used to generate plausible static \threeD human-object interaction configurations from high level textual description. %
\cite{iMapper2018} builds on the \pigraph dataset to define a database of ``scenelets'', %
that are then fitted in \rgb videos to reconstruct plausible dynamic interaction configurations over space-time. %
Finally, \cite{scenegrok2014savva} employs similar observations to predict action maps in a \threeD scene. %
However, these works capture noisy human poses and do not make use of scene constraints to improve them. They also represent human pose as a \threeD skeleton, not a full 3D body. %

Other works like \cite{grabner2011chair,shape2pose2014} use synthetic \threeD scenes and place virtual humans in them to reason about affordances.
\cite{grabner2011chair} do this by using defined key poses of the body
and evaluating human-scene distances and mesh intersections. %
These methods do not actually capture people in scenes.  Our approach could provide rich training data for methods like these to reason about affordances.

{\bf Human \& World Constraints:}
Other works employ human-world interactions more explicitly to establish physical constraints, \ie either contact or collision constraints.
Yamamoto and Yagishita \cite{Yamamoto2000} were the first to use scene constraints in 3D human tracking.
They observed that the scene can constrain the position, velocity and acceleration of an articulated 3D body model.
Later work  adds object contact constraints to the body to effectively reduce the degrees of freedom of the body and make pose estimation easier
\cite{Black_TrackPople,rosenhahn2008}.
Brubaker et al.~\cite{Brubaker2009} focus on walking and perform \threeD person tracking
by using a kinematic model of the torso and the lower body as a prior over human motion and
conditioning its dynamics on the \twoD Anthropomorphic Walker~\cite{kuo2001simple}.
Hasler et al. \cite{5206859} reconstruct a rough \threeD scene from multiple unsynchronized moving cameras and employ scene constraints for pose estimation.
The above methods all had the right idea but required significant manual intervention or were applied in very restricted scenarios.

Most prior methods that have used world constraints focus on interaction with a ground plane \cite{Vondrak2013} or simply constrain the body to move along the ground plane \cite{zhao2004}.
Most interesting among these is the work of Vondrak et al.~\cite{Vondrak2013} where they exploit a game physics engine to infer human pose using gravity, motor forces, and interactions with the ground.
This is a very complicated optimization and it has not been extended beyond ground contact.

Gupta et al.~\cite{gupta2008} exploit contextual scene information in human pose estimation using a GPLVM learning framework.
For an action like sitting, they take motion capture data of people sitting on objects of different heights.
Then, conditioned on the object height, they estimate the pose in the image, exploiting the learned pose model.

Shape2Pose \cite{shape2pose2014} %
learns a model to generate plausible \threeD human poses that interact with a given \threeD object.
First contact points are inferred on the object surface and then the most likely pose that encourages close proximity of relevant body parts to contact points is estimated. %
However, the approach only uses synthetic data.
\cite{zanfir2018monocular} establish contact constraints between the feet and an estimated ground plane.
For this they first estimate human poses in multi-person \rgb videos independently and fit a ground plane around the ankle joint positions.
They then refine poses in a global optimization scheme over all frames incorporating contact and temporal constraints, as well as
collision constraints, using a collision model comprised of shape primitives similar to \cite{bogo2016keep,Oikonomidis_1hand_object}.
More recently, \cite{li2019motionforcesfromvideo} introduced a method to estimate contact positions, forces and torques actuated by the human limbs during human-object interaction.

The \threeD hand-object  community has also explored similar physical constraints, such as \cite{kyriazis2013,Oikonomidis_1hand_object,pham2018pami,Rogez:ICCV:2015,Tsoli:2018:ECCV,Tzionas:IJCV:2016} to name a few. %
Most of these methods employ a collision model to avoid hand-object inter-penetrations with varying degrees of accuracy; using
underlying shape primitives \cite{kyriazis2014,Oikonomidis_1hand_object} or
decomposition in convex parts of more complicated objects \cite{kyriazis2014}, or
using the original mesh to detect colliding triangles along with \threeD distance fields \cite{Tzionas:IJCV:2016}.
Triangle intersection tests have also been used to estimate contact points and forces \cite{Rogez:ICCV:2015}. %
Most other work uses simple proximity checks \cite{sridhar2016objectDexter,Tsoli:2018:ECCV,Tzionas:IJCV:2016} %
and employs an attraction term at contact points. %
Recently, \cite{hasson19_obman} propose an end-to-end model that exploits a contact loss and inter-penetration penalty to reconstruct hands manipulating objects in \rgb images.

In summary, past work focuses either on specific body parts (hands or feet) or interaction with a limited set of objects (ground or hand-held objects).
Here, for the first time, we address the full articulated body interacting with diverse, complex and full \threeD scenes. Moreover, we show how using the 3D scene improves monocular 3D body pose estimation.

\section{Technical Approach}		\label{sec:technical}

\subsection{\threeD Scene Representation}	\label{sec:technical_scene_reconstr}

To study how people interact with a scene, we first need to acquire knowledge about it, \ie to perform scene reconstruction. 
Since physical interaction takes place through surfaces,  %
we chose to represent the scene as a
\threeD mesh $\mesh_\scene = (V_{\scene}, F_{\scene})$, with 
$|V_{\scene}|=N_{\scene}$ vertices $V_{\scene} \in \mathbb{R}^{(N_{\scene} \times 3)}$ and triangular faces $F_{\scene}$. 
We assume a static \threeD scene and reconstruct $\mesh_\scene$ with a standard commercial solution; the \structureIO \cite{structureIO} camera and the \skanect \cite{skanect} software. %
We chose the scene frame to represent the world coordinate frame; both the camera and the human model are expressed \wrt this as explained in Sections \ref{sec:technical_camera} and \ref{sec:technical_body_model}, respectively. 

\subsection{Camera Representation}	\label{sec:technical_camera}

We use a \kinectTWO camera \cite{kinectV2} to acquire 
\rgb and depth 
images of a person moving and interacting with the scene. 
We use a publicly available tool \cite{monocleORIGINAL} 
to estimate the intrinsic camera parameters $K_\cam$ and to capture synchronized \rgbD images; 
for each time frame $t$ we capture a $512 \times 424$ depth image $\imgDDD^t$ and $1920 \times 1080$ RGB image $\imgRGB^t$ at $30$ FPS. We then tranform the \rgbD data into point cloud $P^t$.

To perform human \mocap \wrt to the scene, we first need to register the \rgbD camera to the \threeD scene. 
We assume a static camera and estimate the extrinsic camera parameters, \ie the camera-to-world rigid transformation $T_\cam= (R_\cam,t_\cam)$, where $R_\cam \in SO(3)$ is a rotation matrix and $t_\cam \in \mathbb{R}^3$ is a translation vector. %
For each sequence a human annotator annotates $3$ correspondences between the \threeD scene $\mesh_\scene$ and the  point cloud $P^t$  %
to get an initial estimate of $T_\cam$, which is then refined using ICP \cite{ICP,open3D}.
The camera extrinsic parameters $(R_\cam,t_\cam)$ are fixed during each recording 
(Section \ref{sec:technical_human_mocap_frames}), 

The human body $\body$ is estimated in the camera frame and needs to be registered to the scene by applying $T_\cam$ to it too. 
For simplicity of notation, we use the same symbols for the camera $\cam$ and body $\body$ after transformation to the world coordinate frame. 

\subsection{Human Body Model}	\label{sec:technical_body_model}

We represent the human body using \smplHF \cite{smplifyPP}. %
\smplHF is a generative model that captures how the human body shape varies across a human population, learned from a corpus of registered \threeD body, face and hand scans of people of different sizes, genders and nationalities in various poses. 
It goes beyond similar models \cite{Anguelov05,hasler2009statistical,SMPL:2015,romero2017embodied} by holistically modeling the body with facial expressions and finger articulation, which is important for interactions. %

\smplHF is a differentiable function $\mesh_\body(\beta, \theta, \psi, \trans)$ parameterized by shape $\beta$, pose $\theta$, facial expressions $\psi$ and translation $\trans$. 
Its output is a \threeD mesh $\mesh_\body = (V_{\body}, F_{\body})$ for the human body, with 
$N_{\body}=10475$ vertices $V_{\body} \in \mathbb{R}^{(N_{\body} \times 3)}$ 
and triangular faces $F_{\body}$. 
The shape parameters $\beta \in \mathbb{R}^{10}$ are coefficients in a lower-dimensional shape space learned from approximately 4000 registered \caesar \cite{CAESAR} scans. 
The pose of the body is defined by linear blend skinning with an underlying rigged skeleton, whose \threeD joints $J(\beta)$ are regressed from the mesh vertices.
The skeleton has $55$ joints in total; $22$ for the main body (including a global pelvis joint), %
$3$ for the neck and the two eyes, 
and $15$ joints per hand for finger articulation. 
The pose parameters $\theta=(\theta_\body, \theta_\face, \theta_\hand)$ are comprised of 
$\theta_\body \in \mathbb{R}^{66}$ and $\theta_\face \in \mathbb{R}^{9}$ parameters in axis-angle representation for the main body and face joints respectively, with $3$ degrees of freedom (DOF) per joint, as well as  
$\theta_\hand \in \mathbb{R}^\ncomps$ pose parameters in a lower-dimensional pose space for finger articulation of both hands, captured by approximately 1500 registered hand scans \cite{romero2017embodied}. 
The pose parameters $\theta$ and translation vector $\trans \in \mathbb{R}^3$ define a function that transforms the joints a long the kinematic tree $R_{\theta\trans}$. 
Following the notation of \cite{bogo2016keep} we denote posed joints with $R_{\theta\trans}(J(\beta)_i)$ for each joint $i$. 

\subsection{Human \mocap from Monocular Images}	\label{sec:technical_human_mocap_frames}

To fit \smplHF to single \rgb images we employ \smplifyPP \cite{smplifyPP} and extend it to include human-world interaction constraints to encourage contact and discourage inter-penetrations. 
We name our method {\em \prox} for {\em Proximal Relationships with Object eXclusion}. 
We extend \smplifyPP to \SD, which uses both \rgb and an
additional depth input for more accurate registration of human poses to the \threeD scene. 
We also extend \prox to use \rgbD input instead of \rgb only; we call this configuration \proxD.

Inspired by \cite{smplifyPP}, we formulate fitting \smplHF to monocular images as an optimization problem, where we seek to minimize the objective function 
\begin{align}
	E(\beta,\theta,\psi,\trans,\meshScene) =	&							E_J 					+
												\lambda_{D}				E_D 					+
												\lambda_{\theta_\body}	E_{\theta_\body}		+
  												\lambda_{\theta_\face}	E_{\theta_\face}		+	\nonumber\\	
  											&	\lambda_{\theta_h}		E_{\theta_h}    	 	+		  									%
  		  										\lambda_{\alpha}			E_{\alpha}			+
  												\lambda_{\beta}    		E_{\beta}			+
  												\lambda_{\mExpr}    		E_{\mExpr}			+	\nonumber\\
  											&	\lambda_{\mColl}    		E_{\mColl}			+
  												\lambda_{\contact}   		E_\contact
	\label{eq:objective}
\end{align}
where 
$\theta_\body$,
$\theta_\face$ and 
$\theta_\hand$ 
are the pose vectors for the body, face (neck, jaw) and the two hands respectively, %
$\theta=\{ \theta_\body, \theta_\face, \theta_\hand \}$ is the full set of optimizable pose parameters, 
$\trans$ denotes the body translation, 
$\beta$ the body shape and 
$\psi$ the facial expressions, as described in Section \ref{sec:technical_body_model}. 
$E_J(\beta,\theta,\trans,K,\Jest)$ and $E_D(\beta,\theta,\trans,K,\imgDDD)$ are data terms that are described below; 
$E_J$ is the \rgb data term used in all configurations, while $E_D$ is the optional depth data term which is used whenever depth data is available. 
The terms %
$E_{\theta_\hand}(\theta_\hand)$, 
$E_{\theta_\face}(\theta_\face)$, 
$E_{\mExpr}(\mExpr)$ and 
$E_{\beta}(\beta)$ 
are  $L2$ priors for the hand pose, facial pose, facial expressions and body shape, penalizing deviation from the neutral state. 
Following \cite{bogo2016keep,smplifyPP} the term $E_{\alpha}(\theta_\body) = \sum_{i \in (elbows,knees)}\exp(\theta_i)$  is a prior penalizing extreme bending only for elbows and knees, while 
$E_{\theta_\body}(\theta_\body)$ is a VAE-based body pose prior called VPoser introduced in \cite{smplifyPP}. 
The term $E_\contact(\beta,\theta,\trans,\meshScene)$ encourages contact between 
the body and the scene as described in Section \ref{sec:technical_contact}. 
The term $E_{\mColl}(\theta, \beta, \meshScene)$ is a penetration penalty modified from \cite{smplifyPP} to reason about both self-penetrations and human-scene inter-penetrations, 
as described in Section \ref{sec:technical_collisions}. 
The terms $E_J$, $E_{\theta_\body}$, 
$E_{\theta_\hand}$, 
$E_{\alpha}$, $E_{\beta}$ and weights $\lambda_i$ are as described in \cite{smplifyPP}. 
The weights $\lambda_i$ denote steering weights for each term. They were set empirically in an annealing scheme similar to \cite{smplifyPP}. 

For the {\it \rgb data term} $E_J$ we use a re-projection loss to minimize the weighted robust distance between \twoD joints $\Jest(\imgRGB)$ estimated from the \rgb image $\imgRGB$ and the \twoD projection of the corresponding posed \threeD joints 
$R_{\theta\trans}(J(\beta)_i)$
of \smplHF, as defined for each joint $i$ in Section \ref{sec:technical_body_model}. 
Following the notation of \cite{bogo2016keep,smplifyPP}, 
the data term is %
\begin{eqnarray}
	\lefteqn{E_J(\beta,\theta,\trans,K,\Jest) =}\nonumber\\
& &	\sum_{joint~i} \annealingGamma_i \omega_i \rho_J(   \mathit{\Pi}_K(   R_{\theta\trans}(J(\beta)_i)  -   J_{est,i}   )
	\label{eq:data_term}
\end{eqnarray}
where $\mathit{\Pi}_K$ denotes the \threeD to \twoD projection with intrinsic camera parameters $K$. 
For the \twoD detections we rely on \openpose \cite{cao2017realtime,simon2017hand,wei2016convolutional}, which provides body, face and hands keypoints jointly for each person in an image.
To account for noise in the detections, the contribution of each joint in the data term is weighted by the detection confidence score $\omega_i$, while $\annealingGamma_i$ are per-joint weights for annealed optimization, as described in \cite{smplifyPP}. 
Furthermore, $\rho_J$ denotes a robust \mbox{Geman-McClure} error function \cite{GemanMcClure1987} for down-weighting noisy detections.

The {\it depth data term} $E_D$ minimizes the discrepancy between the visible body vertices $V_{\body}^\visible \subset V_{\body}$ and a segmented point cloud $P^t$
that belongs only to the body and not the static scene.  For this, we
use the body segmentation mask 
from the \kinectTWO SDK. 
Then, $E_D$ is defined as
\begin{equation}
E_D(\beta,\theta,\trans,K,\imgDDD)   =   \sum_{p \in P^t}   \rho_D(   \min_{v \in V_{\body}^\visible}   \| v - p \|   )
\end{equation}
where $\rho_D$ denotes a robust \mbox{Geman-McClure} error function \cite{GemanMcClure1987} for downweighting vertices $V_{\body}^\visible$ that are far from $P^t$. 

\subsection{Contact Term}	\label{sec:technical_contact}

\begin{figure}
    \centering                      %
    \subfloat{	\includegraphics[trim=000mm 000mm 000mm 000mm, clip=false, width=0.50 \linewidth]{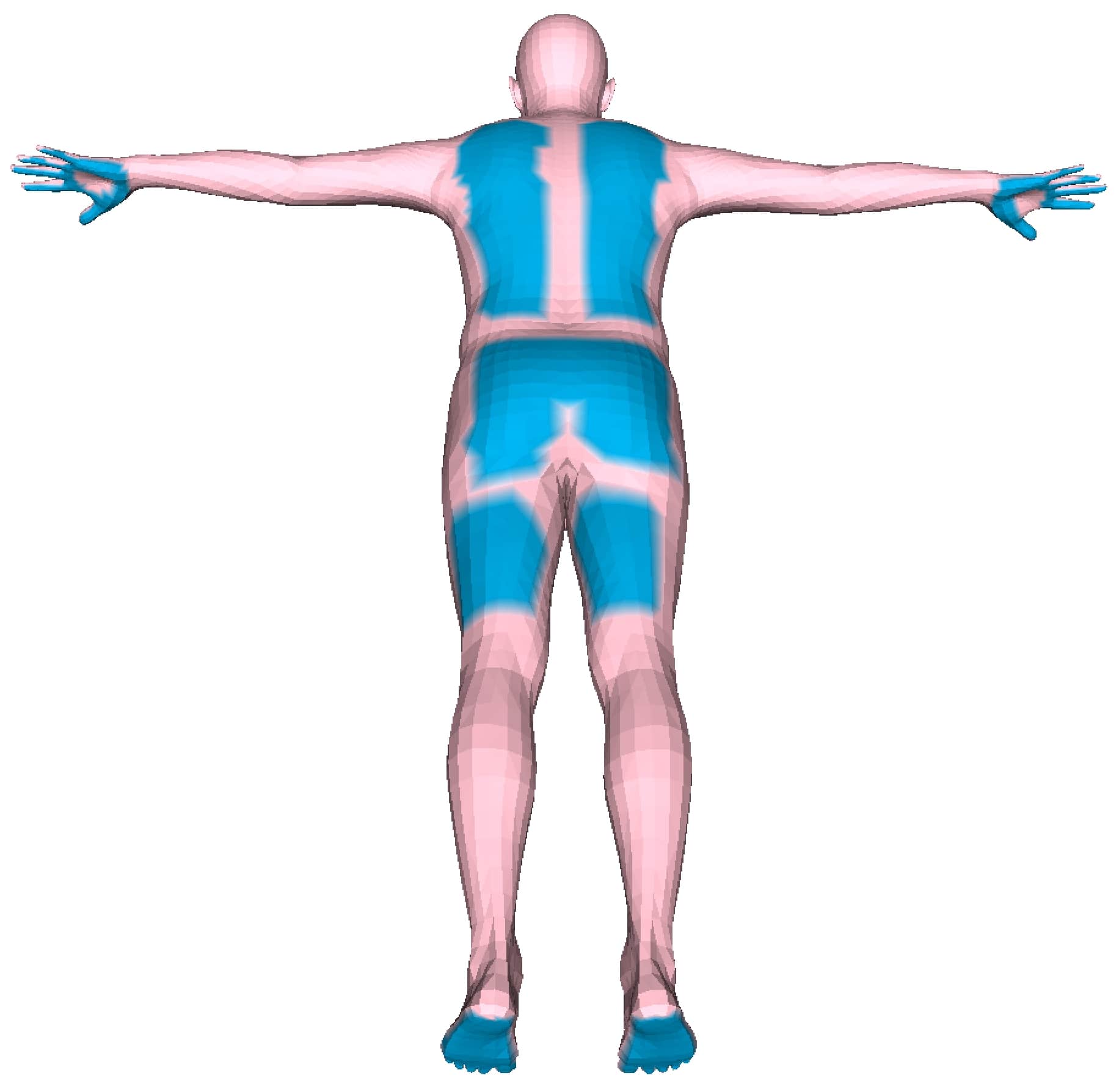}	}
    \vspace*{-04.00mm}
    \caption{
    				Annotated vertices that come frequently in contact with the world, highlighted with blue color. 
    	}
    \label{fig:contact_vertices}
\end{figure}

Using 
the \rgb term $E_J$ without reasoning about human-world interaction %
might result in physically implausible poses, as shown in Figure \ref{fig:teaser}; 
However, when humans interact with the scene they come in \emph{contact} with it, \eg feet contact the floor while standing or walking. 
We therefore introduce the term $E_\contact$ to 
encourage contact and \emph{proximity} between body parts and the scene around contact areas. 

To that end, we annotate  
a set of candidate contact vertices $V_\contact \subset V_\body$ across the whole body that come frequently in contact with the world, focusing on the actions of walking, sitting and touching with hands. 
We annotate
$1121$ vertices across the whole body, as shown in Figure \ref{fig:contact_vertices}. 
We also explored choosing all body vertices as contact vertices but found that this choice is suboptimal, for evaluation see \supmat~
We define the contact vertices as:
$725$  vertices for the hands, 
$62$   vertices for the thighs, 
$113$  for the gluteus, %
$222$  for the back, and 
$194$  for the feet. $E_\contact$ is defined as:
\begin{equation}
E_\contact(\beta,\theta,\trans,\meshScene)   =   \sum_{v_\contact \in V_{\contact}}   \rho_\contact(   \min_{v_s \in V_\scene}   \| v_\contact - v_s \|   )
\end{equation}
where 
$\rho_\contact$ denotes a robust \mbox{Geman-McClure} error function
\cite{GemanMcClure1987} for down-weighting vertices in $V_\contact$
that are far from the nearest vertices in $V_\scene$ of the \threeD scene $\mesh_\scene$. 

\subsection{Penetration Term}	\label{sec:technical_collisions}

Intuitive physics suggests that two objects can not share the same \threeD space. 
However, human pose estimation methods might result in self-penetrations or bodies penetrating surrounding \threeD objects, 
as shown in Figure \ref{fig:teaser}. %
We therefore introduce a penetration term that combines $E_{\mColl_\colisSelfTR}$ and $E_{\mColl_\colisInterTR}$ that are defined below:
\begin{eqnarray}\label{eq:collisionDetection}
\lefteqn{	E_{\mColl}(\theta,\beta,\trans,\meshScene) = }\nonumber \\
 & & E_{\mColl_\colisSelfTR}( \theta,\beta) +  E_{\mColl_\colisInterTR}(\theta,\beta,\trans, \meshScene)
\end{eqnarray}
For \emph{self-penetrations} we follow the approach of \cite{LucaHands,smplifyPP,Tzionas:IJCV:2016}, that follows local reasoning. 
We first detect a list of colliding body triangles 
$\mColl_\colisSelfTR$ 
using Bounding Volume Hierarchies (BVH) \cite{collisionDeformableObjects}
and compute local conic \threeD distance fields $\Psi$.
Penetrations are then penalized according to the depth in $\Psi$.
For the exact definition of $\Psi$ and $ E_{\mColl_\colisSelfTR}(\para,\beta)$ we refer the reader to \cite{LucaHands,Tzionas:IJCV:2016}. 

For body-scene \emph{inter-penetrations} local reasoning at colliding triangles is not enough, as the body might be initialized deep inside \threeD objects or even outside the \threeD scene. 
To resolve this, we penalize all penetrating vertices using the signed distance field (SDF) of the scene $\meshScene$. 
The distance field is represented with a uniform voxel grid with size $256 \times 256 \times 256$, that spans a padded bounding box of the scene. 
Each voxel cell $c_i$ stores the distance from its center 
$p_i \in \mathbb{R}^3$ 
to the nearest surface point $p_i^s \in \mathbb{R}^3$ of $\meshScene$ with normal $n_i^s \in \mathbb{R}^3$, while 
the sign is defined %
according to the relative orientation of the vector $p_i - p_i^s$ \wrt $n_i^s$
as 
\begin{equation}
	\label{eq:df_sign}
	sign\left(c_i\right) = sign\left(\left(     p_i  -      p_i^s \right)\cdot      n_i^s  \right) ;
\end{equation}
a positive sign means that the body vertex is outside the nearest
scene object, while a negative sign means that it is inside the nearest scene object and denotes penetration. 
In practice, during optimization we can find how each body vertex $V_{\body_i}$ is positioned relative to the scene by reading the signed distance $d_i \in \mathbb{R}$ of the voxel it falls into. 
Since the limited grid resolution influences discretization of the \threeD distance field, we perform trilinear interpolation using the neighboring voxels similar to \cite{stn2015}. 
Then we resolve body-scene inter-penetration by minimizing the loss term 
\begin{equation}
    \label{eq:body_scene_loss}
    E_{\mColl_\colisInterTR} = \sum_{d_i < 0} \Vert d_i      n_i^s  \Vert^2 .
\end{equation}
\subsection{Optimization}	\label{sec:technical_optimization}

We optimize Equation \ref{eq:objective} similar to \cite{smplifyPP}. 
More specifically, we implement our model in \pytorch and use the Limited-memory BFGS optimizer (L-BFGS) \cite{nocedal2006nonlinear} with strong Wolfe line search. %

\section{Datasets}				\label{sec:datasets}

\begin{figure}
    \centering                      %
    \subfloat{	\includegraphics[trim=000mm 000mm 000mm 000mm, clip=false, width=0.98 \linewidth]{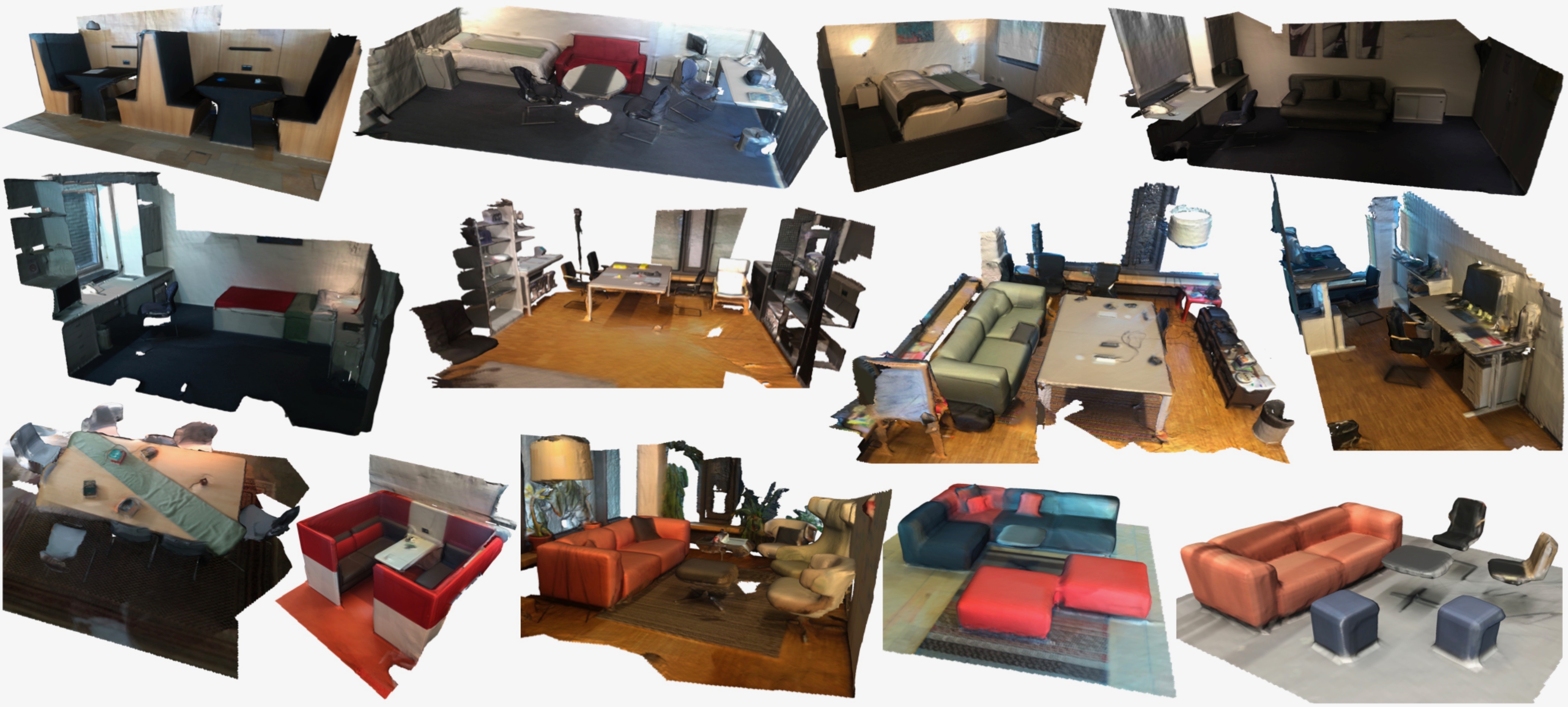}	}
    \vspace*{-02.00mm}
    \caption{
    							Reconstructed \threeD scans of the $12$ indoor scenes of our \prox dataset, as well as an additional scene for our quantitative dataset, shown at the bottom right corner.
	}
    \label{fig:our_dataset_ReconstructedRooms}
\end{figure}

\begin{figure}[t]
    \centering                      %
    \subfloat{	\includegraphics[trim=000mm 000mm 10mm 000mm, clip=true, width=0.98 \linewidth]{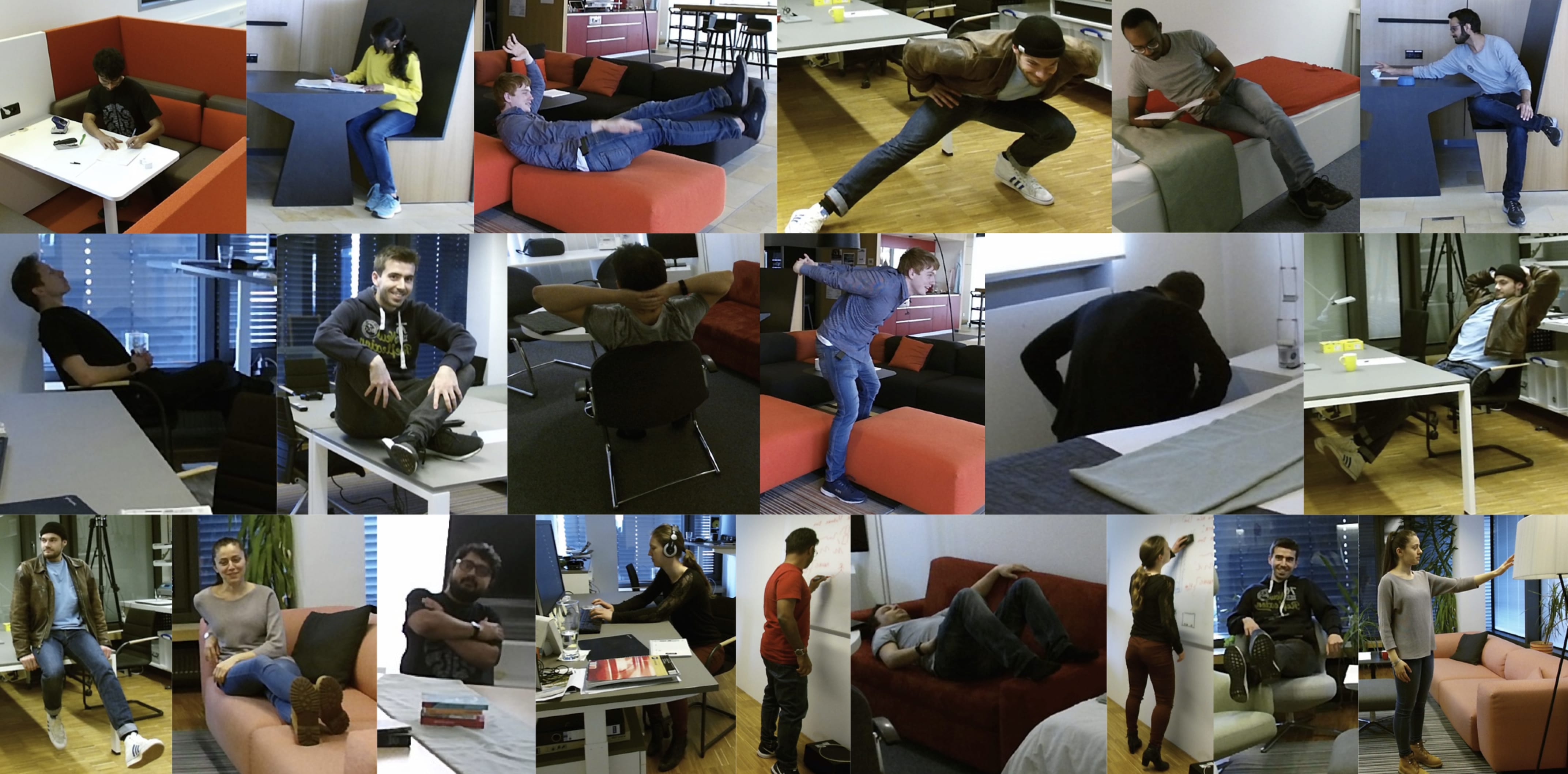}	}
    	\vspace*{-02.00mm}
    \captionof{figure}{
    							Example \rgb frames of our \prox dataset showing people moving in natural indoor scenes and interacting with them. 
    							We reconstruct in total $12$ scenes and capture $20$ subjects. 
    							Figure \ref{fig:our_dataset_ReconstructedRooms} shows the \threeD reconstructions of our indoor scenes. 
    	}
    \label{fig:our_dataset_RGB_samples}
\end{figure}

\subsection{Qualitative Datasets}		\label{sec:dataset___our_big}
The qualitative datasets, \pigraph and \prox, contain: \threeD scene scans and monocular videos of people interacting with the \threeD scenes. They do not include \gt bodies, thus we cannot evaluate our method quantitatively on these datasets. 
\subsubsection{\pigraph dataset}	
This dataset was released as part of the work of Sava \emph{et al.} \cite{savva2016pigraphs}. The dataset has several \threeD scene scans and \rgbD videos. 
It suffers from multiple limitations; 
the color and depth frames are neither synchronized nor spatially calibrated,  making it hard to use both \rgb and depth. 
The human poses are rather noisy and are not well registered into the \threeD scenes, %
which are inaccurately reconstructed. %
The dataset has a low frame rate of $5$ fps, 
it is limited to only $5$ subjects and does not have \gt. 

\subsubsection{\prox dataset}	
We collected this dataset to overcome the limitations of the \pigraph dataset.
We employ the commercial \structureIO \cite{structureIO} \rgbD camera and the accompanying \threeD reconstruction solution \skanect \cite{skanect} and reconstruct \ourDataSCENES indoor scenes, shown in Figure \ref{fig:our_dataset_ReconstructedRooms}. 
The scenes can be grouped to: $3$ bedrooms, $5$ living rooms, $2$ sitting booths and $2$ offices.
We then employ a \kinectTWO \cite{kinectV2} \rgbD camera to capture \ourDataSUBJECTS subjects ($4$ females and $16$ males) interacting with these scenes. 
Subjects gave written informed consent to make their data available for research purposes.
The dataset provides $100$K synchronized and spatially calibrated \rgbD frames at $30$ fps. 
Figure \ref{fig:our_dataset_RGB_samples} shows example \rgb frames from our dataset. %
We leverage the \rgbD videos to get pseudo \gt by extending \smplifyPP to \SD which fits \smplHF to both \rgb and depth data instead of \rgb only.

\subsection{Quantitative Dataset}			\label{sec:dataset___our_gt}
Neither our \prox dataset %
nor \pigraph \cite{savva2016pigraphs} have \gt for quantitative evaluation. 
To account for this, we captured a separate \emph{quantitative dataset} with \ourViconFrames static \rgbD frames in sync with a $54$ camera \vicon system.
We placed markers on the body and the fingers. 
We placed everyday furniture and objects inside the \vicon area to mimic a living room, %
and performed \threeD reconstruction of the scene, shown in the bottom right corner of Figure \ref{fig:our_dataset_ReconstructedRooms} with the \structureIO \cite{structureIO} and \skanect \cite{skanect} similar to above. %
We then use \moshPP \citeMOSH which is a method that converts \mocap data into realistic 3D human meshes represented by a rigged body model.
Example \rgb frames are shown in Figure \ref{fig:mosh_dataset___rgb_gt_results} (left), while our mesh pseudo \gt is shown with aqua blue color. 

Our datasets will be available for research purposes.

\section{Experiments}			\label{sec:experiments}
\begin{table}
	\begin{center}
	\footnotesize
	\setlength{\tabcolsep}{1pt}
	\begin{tabular}{lc|c|c|c|c|c|c|c|c|c|c|c|}																																																																																																												\hhline{~~----~----~~}
	\multicolumn{1}{c}{}	&	\multicolumn{1}{c|}{}	&	\multicolumn{4}{c|}{~Eq. \ref{eq:objective} terms~}																																														&	\multicolumn{1}{c|}{}	&	\multicolumn{4}{c|}{Error}																				& \multicolumn{1}{c}{} 	& \multicolumn{1}{c}{} 	\\	\hhline{~~----~----~~}
	\noalign{\smallskip}																																																																																																																		\hhline{~~----~----~~}
	\multicolumn{1}{c}{}	&	\multicolumn{1}{c|}{}	&	\multirow{1}{*}{~$E_J$~}								&		\multirow{1}{*}{~$E_\contact$~}							&	\multirow{1}{*}{~$E_{\mColl}$~}							&	\multirow{1}{*}{~$E_D$~}								&	\multicolumn{1}{c|}{}	&	~PJE~						&	~V2V~					&	~p.PJE~				&	~p.V2V~				& \multicolumn{1}{c}{} 	& \multicolumn{1}{c}{} 	\\	\hhline{~~----~----~~}
	\noalign{\smallskip}																																																																																																																		\hhline{~~----~----~-}
	\multirow{6}{*}{~(a)~}							& {} & {~\cmark~} & {~\xmark~} & {~\xmark~} & {~\xmark~} & {}	&	~$220.27$~	&	~$218.06$~	&	~$73.24$~	&	~$60.80$~ 	&	{}	&	\multirow{6}{*}{\centering\begin{turn}{90}mm\end{turn}}																																																										\\	\hhline{~~----~----~~}	
	{}												& {} & {~\cmark~} & {~\cmark~} & {~\xmark~} & {~\xmark~} & {}	&	~$208.03$~	&	~$208.57$~	&	~$72.76$~	&	~$60.95$~ 	&	{}	&	{}																																																																							\\	\hhline{~~----~----~~}
	{}												& {} & {~\cmark~} & {~\xmark~} & {~\cmark~} & {~\xmark~} & {}	&	~$190.07$~	&	~$190.38$~	&	~$73.73$	~	&	~$62.38$~ 	&	{}	&	{}																																																																							\\	\hhline{~~----~----~~}
	{}												& {} & {~\cmark~} & {~\cmark~} & {~\cmark~} & {~\xmark~} & {}	&	~$\textbf{167.08}$~	&	~$\textbf{166.51}$~	&	~$\textbf{71.97}$	~	&	~$\textbf{61.14}$~ 	&	{}	&	{}																																																																							\\	\hhline{~~----~----~~}
	{}												& {} & {~\cmark~} & {~\xmark~} & {~\xmark~} & {~\cmark~} & {}	&	~$72.91$~	&	~$69.89$~	&	~$55.53$	~	&	~$48.86$~ 	&	{}	&	
	{}																															\\	\hhline{~~----~----~~}					
	{}												& {} & {~\cmark~} & {~\cmark~} & {~\cmark~} & {~\cmark~} & {}	&	~$\textbf{68.48}$~	&	~$\textbf{60.83}$~	&	~$\textbf{52.78}$	~	&	~$\textbf{47.11}$~ 	&	{}	&	{}	
	\\	\hhline{~~----~----~-}
	\noalign{\smallskip}																																																																																																																		\hhline{~~----~----~-}
	\multirow{2}{*}{~(b)~}							& {} & {~\cmark~} & {~\xmark~} & {~\xmark~} & {~\xmark~} & {}	&	~$232.29$~	&	~$227.49$~	&	~$66.02$~	&	~$53.15$~ 	&	{}	&	\multirow{2}{*}{\centering\begin{turn}{90}mm\end{turn}}																																																										\\	\hhline{~~----~----~~}	
	{}												& {} & {~\cmark~} & {~\cmark~} & {~\cmark~} & {~\xmark~} & {}	&	~$\textbf{144.60}$~	&	~$\textbf{156.90}$~	&	~$\textbf{65.04}$	~	&	~$\textbf{52.60}$~ 	&	{}	&	{}																																																																							\\	\hhline{~~----~----~-}
	\end{tabular}
	\end{center}
    \vspace{-05.00mm}
	\caption{
						Ablation study for Equation \ref{eq:objective}; each row contains the terms indicated by the check-boxes.  Units in $mm$. \prox and \proxD are shown in bold.
						{\bf Table (a):} 
						Evaluation on our \emph{quantitative dataset} using mesh pseudo \gt based on \vicon and \moshPP \citeMOSH. 
						{\bf Table (b):} 
						Evaluation on     chosen sequences of our \emph{qualitative dataset} using pseudo \gt based on \SD. 
						{\bf Tables (a, b):} 
						We report 
						the mean per-joint error without/with procrustes alignment noted as ``PJE'' / ``p.PJE'', and %
						the mean vertex-to-vertex error noted %
														as ``V2V'' / ``p.V2V''.
	}
	\label{tab:ablation}
\end{table}

{\bf Quantitative Evaluation:} 
To evaluate the performance of our method, as well as to evaluate the importance of different terms in Equation \ref{eq:objective}, we perform quantitative evaluation in Table \ref{tab:ablation}. 
As performance metrics we report 
the mean per-joint error without and with procrustes alignment noted as ``PJE'' and ``p.PJE'' respectively, as well as 
the mean vertex-to-vertex error noted similarly as ``V2V'' and ``p.V2V''. 
Each row in the table shows a setup that includes different terms as indicated by the check-boxes. 
Table \ref{tab:ablation} includes two sub-tables for different datasets. %
\noindent
{\bf \mbox{Table 1 (a)}:}
We employ our new \emph{quantitative dataset} with mesh pseudo \gt based on \vicon and \moshPP \citeMOSH, as described in Section \ref{sec:datasets}. 
The first row with only $E_J$ is an \rgb-only  baseline similar to \smplifyPP \cite{smplifyPP}, that we adapt to our needs by using a fixed camera and estimating body translation $\trans$, and gives the biggest ``PJE'' and ``V2V'' error. %
In the second row we add only the contact term $E_\contact$, while in the third row we add only the penetration term $E_\mColl$. 
In both cases the error drops a bit, however the drop is significantly bigger for the fourth row that includes both $E_\contact$ and $E_\mColl$; this corresponds to \emph{\prox} and achieves $167.08$ mm ``PJE'' and $166.51$ mm ``V2V'' error. 
This suggests that both $E_\contact$ and $E_\mColl$ contribute to accuracy and are complementary. 
To inform the upper bound of performance, in the fifth row we employ an \rgbD baseline with $E_J$ and $E_D$, which corresponds to \SD as described in Section \ref{sec:technical_human_mocap_frames}. All terms of Equation \ref{eq:objective} are employed in the last row; we call this configuration \proxD. We observe that using scene constraints boosts the performance even when the depth is available. 
This gives the best overall performance, but \prox (fourth row) achieves reasonably good performance with less input data, \ie using \rgb only. 
\noindent
{\bf \mbox{Table 1 (b)}:}
We chose $4$ random sequences of our new \prox dataset. We generate pseudo \gt with \SD, which uses both \rgb and depth. 
We show a comparison between the \rgb-only baseline (first row) and \prox (second row) compared to the pseudo \gt of \SD. 
The results support the above finding %
that the scene constraints in \prox contribute significantly to accuracy. 

The run time for all configurations is reported in the \supmat

\newcommand{\moshFigureSIZ}{0.10}
\newcommand{\moshFigureVERT}{-04.00mm}
\newcommand{\moshFigureHRZa}{-02.40mm}

\begin{figure*}
    \centering
    \subfloat{	\includegraphics[trim=000mm 000mm 000mm 000mm, clip=false,
    width=0.98\linewidth]{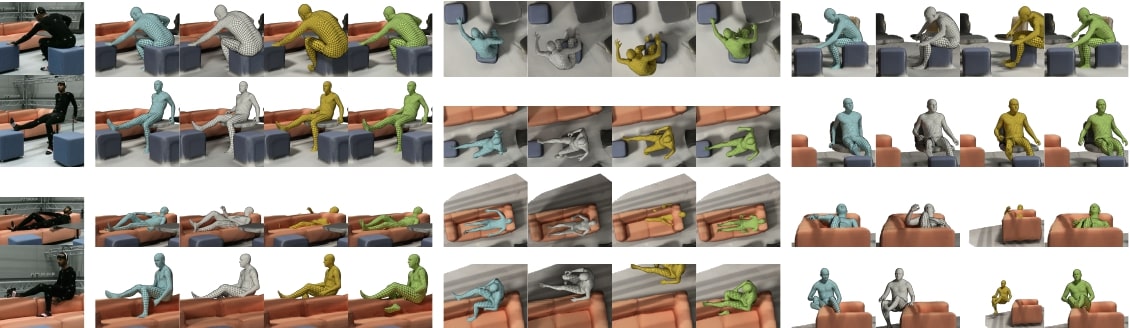}	}
    \caption{%
        Examples from our \emph{quantitative dataset}, described in Section~\ref{sec:experiments}.
        From left to right: 
        (1) RGB images, 
        (2) rendering of the fitted model and the \threeD scene from the camera viewpoint; aqua blue for the mesh pseudo \gt, light gray for the results of our method \prox, yellow for results without scene constraints, green for \SD, 
        (3) top view and 
        (4) side view. 
        More results can be found in \supmat
    }
    \label{fig:mosh_dataset___rgb_gt_results}
\end{figure*}

\begin{figure*}
    \centering                      %
    \subfloat{	\includegraphics[trim=000mm 000mm 000mm 000mm, clip=false, width=0.97\linewidth]{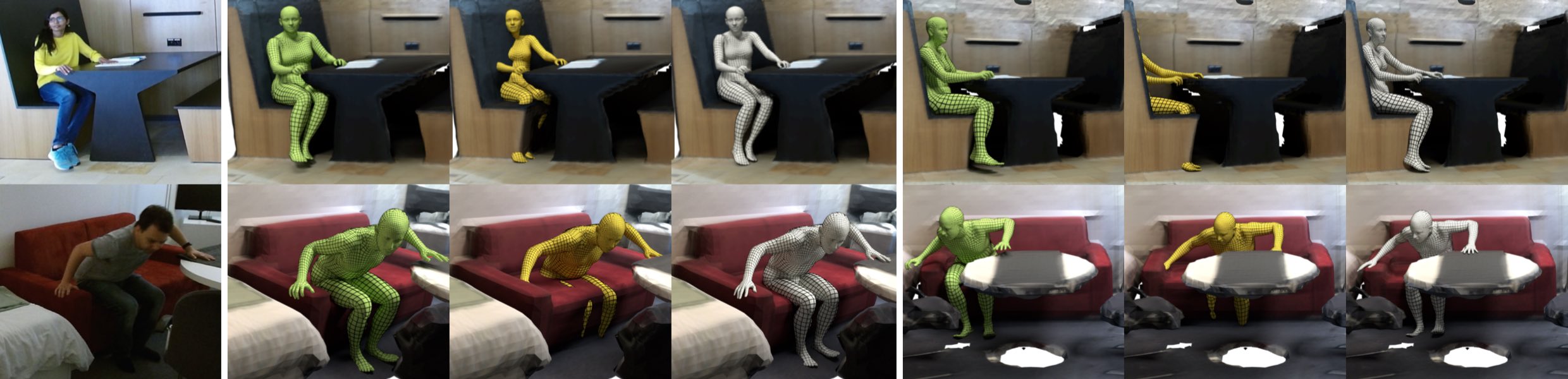}	}    \\	\vspace{-03.00mm}
    \subfloat{	\includegraphics[trim=000mm 000mm 000mm 000mm, clip=false, width=0.97\linewidth]{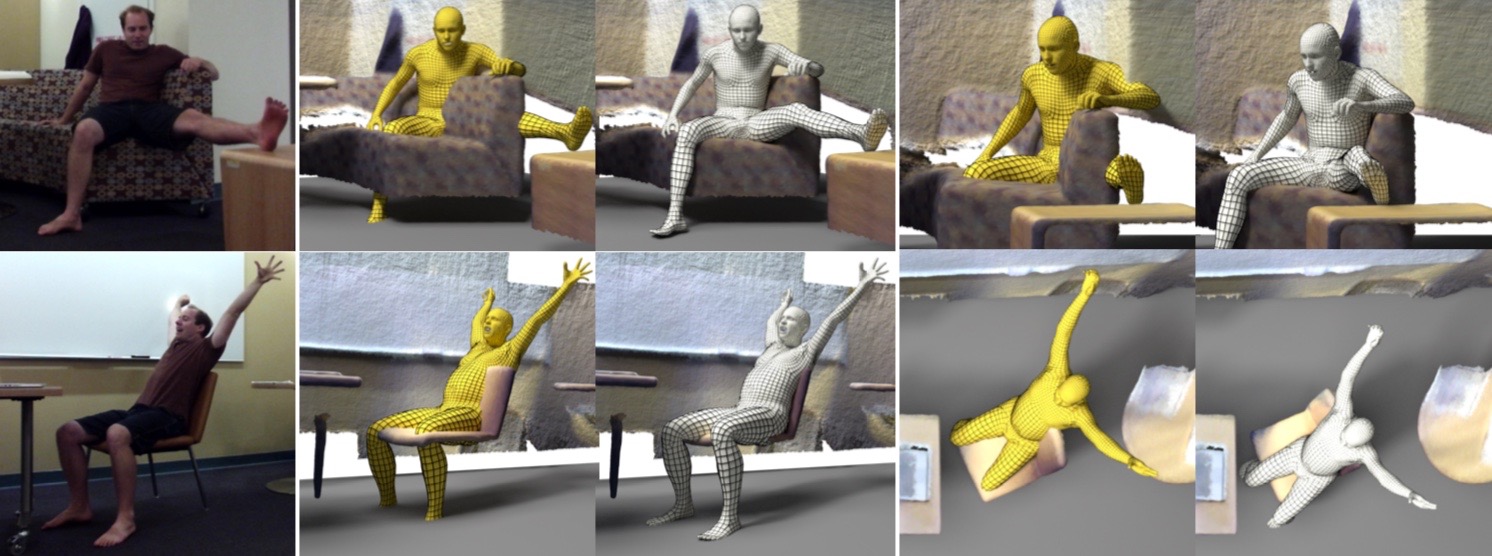}	}
    \caption{
        Qualitative results of our method on two datasets; on our \emph{qualitative dataset} (top set) and on the \pigraph dataset \cite{savva2016pigraphs} (bottom set).
        From left to right: 
        (1) RGB images, 
        (2) rendering from the camera viewpoint; light gray for the results of our method \prox, yellow for results without scene constraints, and green for \SD (applicable only for the top set), 
        (3) rendering from a different view, that shows that the camera view is deceiving.
        More results can be found in \supmat
    }
    \label{fig:results___our_method_many}
    \vspace{-1em}
\end{figure*}
{\bf Qualitative Evaluation:} 
In Figure \ref{fig:mosh_dataset___rgb_gt_results} we show qualitative results too for our \emph{quantitative dataset}. 
Furthermore, in Figure \ref{fig:results___our_method_many} we show representative qualitative results on the \emph{qualitative datasets}; our \prox dataset and \pigraph dataset. 
In both figures, the lack of scene constraints (yellow) results in severe penetrations in the scene. 
Our method, \prox, includes scene constraints (light gray) and estimates bodies that are significantly more consistent with the \threeD scene, \ie with realistic contact and without penetrations. More qualitative results are available in the \supmat

\section{Conclusion}				\label{sec:conclusion}

In this work we focus on human-world interactions and capture the motion of humans interacting with a real static \threeD scene in \rgb images.
We use a holistic model, \mbox{\smplHF} \cite{smplifyPP}, that jointly models the body with face and fingers, which are important for interactions.
We show that incorporating interaction-based human-world constraints in an optimization framework (\prox) results in significantly more realistic and accurate \mocap.
We also collect a new dataset of \threeD scenes with RGB-D sequences involving human interactions and occlusions.
We perform extensive quantitative and qualitative evaluations that clearly show the benefits of incorporating scene constraints into \threeD human pose estimation.
Our code, data and \mocap are available for research purposes.

{\bf Limitations and Future work:}
A limitation of the current formulation is that we do not model scene occlusion.
Current \twoD part detectors do not indicate when joints are occluded and may provide inaccurate results.
By knowing the scene structure we could reason about what is visible and what is not.
Another interesting direction would be the unification of the self-penetration and the body-scene inter-penetration
by employing the implicit formulation of~\cite{Taylor:2017:ADF:3130800.3130853} for the whole body.
Future work can exploit recent deep networks to estimate the scene directly from monocular \rgb images. More interesting directions would be to extend our method to dynamic scenes \cite{ruenz2017cofusion}, human-human interaction and to account for scene and body deformation.

{\bf Acknowledgments:}
We thank Dorotea Lleshaj, Markus H{\"o}schle, Mason Landry, Andrea Keller and Tsvetelina Alexiadis for their help with the data collection. Jean-Claude Passy for helping with the data collection software. Nima Ghorbani for \moshPP. Benjamin Pellkofer for the IT support. Jonathan Williams for managing the website.

{\bf Disclosure:}
 MJB has received research gift funds from Intel, Nvidia,
 Adobe, Facebook, and Amazon. While MJB is a part-time employee of
Amazon, his research was performed solely at, and funded solely by, MPI.
MJB has financial interests in Amazon and Meshcapade GmbH.


{\small
\bibliographystyle{ieee_fullname}
\bibliography{egbib,bib_bodies,bib_handrefs,bib_faces,bib_vposer,bib_homogenus}

\begin{thebibliography}{10}\itemsep=-1pt

\bibitem{kinectV2}
Kinect for xbox one.
\newblock
  \url{https://en.wikipedia.org/wiki/Kinect\#Kinect_for_Xbox_One_(2013)}.

\bibitem{monocleORIGINAL}
Monocle: Kinect data capture app.
\newblock \url{https://github.com/bmabey/monocle}.

\bibitem{skanect}
Skanect: 3d scanning.
\newblock \url{https://skanect.occipital.com}.

\bibitem{structureIO}
Structure sensor: 3d scanning, augmented reality and more.
\newblock \url{https://structure.io/structure-sensor}.

\bibitem{aksoy2010categorizing}
Eren~Erdal Aksoy, Alexey Abramov, Florentin W{\"o}rg{\"o}tter, and Babette
  Dellen.
\newblock Categorizing object-action relations from semantic scene graphs.
\newblock In {\em 2010 IEEE International Conference on Robotics and Automation
  (ICRA)}, pages 398--405, 2010.

\bibitem{Anguelov05}
Dragomir Anguelov, Praveen Srinivasan, Daphne Koller, Sebastian Thrun, Jim
  Rodgers, and James Davis.
\newblock {SCAPE: Shape Completion and Animation of PEople}.
\newblock {\em ACM Transactions on Graphics (TOG), (Proc. SIGGRAPH)},
  24(3):408--416, 2005.

\bibitem{armeni20163d}
Iro Armeni, Ozan Sener, Amir~R Zamir, Helen Jiang, Ioannis Brilakis, Martin
  Fischer, and Silvio Savarese.
\newblock 3d semantic parsing of large-scale indoor spaces.
\newblock In {\em The IEEE Conference on Computer Vision and Pattern
  Recognition (CVPR)}, pages 1534--1543, 2016.

\bibitem{LucaHands}
Luca Ballan, Aparna Taneja, Juergen Gall, Luc Van~Gool, and Marc Pollefeys.
\newblock Motion capture of hands in action using discriminative salient
  points.
\newblock In {\em The European Conference on Computer Vision (ECCV)}, pages
  640--653, 2012.

\bibitem{ICP}
Paul~J. Besl and Neil~D. McKay.
\newblock A method for registration of 3-d shapes.
\newblock {\em IEEE Transactions on Pattern Analysis and Machine Intelligence
  (TPAMI)}, 14(2):239--256, 1992.

\bibitem{bogo2016keep}
Federica Bogo, Angjoo Kanazawa, Christoph Lassner, Peter Gehler, Javier Romero,
  and Michael~J Black.
\newblock Keep it {SMPL}: Automatic estimation of 3{D} human pose and shape
  from a single image.
\newblock In {\em The European Conference on Computer Vision (ECCV)}, 2016.

\bibitem{Brubaker2009}
Marcus~A. Brubaker, David~J. Fleet, and Aaron Hertzmann.
\newblock Physics-based person tracking using the anthropomorphic walker.
\newblock {\em International Journal of Computer Vision}, 87(1):140, Aug 2009.

\bibitem{cao2017realtime}
Zhe Cao, Tomas Simon, Shih-En Wei, and Yaser Sheikh.
\newblock Realtime multi-person 2{D} pose estimation using part affinity
  fields.
\newblock In {\em The IEEE Conference on Computer Vision and Pattern
  Recognition (CVPR)}, 2017.

\bibitem{dai2017scannet}
Angela Dai, Angel~X. Chang, Manolis Savva, Maciej Halber, Thomas Funkhouser,
  and Matthias Nie{\ss}ner.
\newblock Scannet: Richly-annotated 3d reconstructions of indoor scenes.
\newblock In {\em The IEEE Conference on Computer Vision and Pattern
  Recognition (CVPR)}, 2017.

\bibitem{delaitre2012scene}
Vincent Delaitre, David~F Fouhey, Ivan Laptev, Josef Sivic, Abhinav Gupta, and
  Alexei~A Efros.
\newblock Scene semantics from long-term observation of people.
\newblock In {\em The European Conference on Computer Vision (ECCV)}, pages
  284--298, 2012.

\bibitem{Eigen:2014}
David Eigen, Christian Puhrsch, and Rob Fergus.
\newblock Depth map prediction from a single image using a multi-scale deep
  network.
\newblock In {\em Advances in Neural Information Processing Systems}, pages
  2366–--2374, 2014.

\bibitem{fisher2015activity}
Matthew Fisher, Manolis Savva, Yangyan Li, Pat Hanrahan, and Matthias
  Nie{\ss}ner.
\newblock Activity-centric scene synthesis for functional 3d scene modeling.
\newblock {\em ACM Transactions on Graphics (TOG)}, 34(6):179, 2015.

\bibitem{fouhey2014people}
David~F Fouhey, Vincent Delaitre, Abhinav Gupta, Alexei~A Efros, Ivan Laptev,
  and Josef Sivic.
\newblock People watching: Human actions as a cue for single view geometry.
\newblock {\em International Journal of Computer Vision (IJCV)},
  110(3):259--274, 2014.

\bibitem{Review_Gavrila}
Dariu~M. Gavrila.
\newblock The visual analysis of human movement: A survey.
\newblock {\em Computer Vision and Image Understanding (CVIU)}, 73(1):82 -- 98,
  1999.

\bibitem{GemanMcClure1987}
Stuart Geman and Donald~E. McClure.
\newblock Statistical methods for tomographic image reconstruction.
\newblock In {\em Proceedings of the 46th Session of the International
  Statistical Institute, Bulletin of the ISI}, volume~52, 1987.

\bibitem{gibson1950perception}
James~J Gibson.
\newblock {\em The perception of the visual world}.
\newblock Houghton Mifflin, 1950.

\bibitem{grabner2011chair}
Helmut Grabner, Juergen Gall, and Luc Van~Gool.
\newblock What makes a chair a chair?
\newblock In {\em The IEEE Conference on Computer Vision and Pattern
  Recognition (CVPR)}, pages 1529--1536, 2011.

\bibitem{gupta2008}
Abhinav Gupta, Trista Chen, Francine Chen, Don Kimber, and Larry~S Davis.
\newblock Context and observation driven latent variable model for human pose
  estimation.
\newblock In {\em The IEEE Conference on Computer Vision and Pattern
  Recognition (CVPR)}, pages 1 -- 8, 2008.

\bibitem{gupta2009observing}
Abhinav Gupta, Aniruddha Kembhavi, and Larry~S Davis.
\newblock Observing human-object interactions: Using spatial and functional
  compatibility for recognition.
\newblock {\em IEEE Transactions on Pattern Analysis and Machine Intelligence
  (TPAMI)}, 31(10):1775--1789, 2009.

\bibitem{gupta2011workspace}
Abhinav Gupta, Scott Satkin, Alexei~A Efros, and Martial Hebert.
\newblock From 3d scene geometry to human workspace.
\newblock In {\em The IEEE Conference on Computer Vision and Pattern
  Recognition (CVPR)}, pages 1961--1968, 2011.

\bibitem{5206859}
Nils Hasler, Bodo Rosenhahn, Thorsten Thormahlen, Michael Wand, J{\"u}rgen
  Gall, and Hans-Peter Seidel.
\newblock Markerless motion capture with unsynchronized moving cameras.
\newblock In {\em The IEEE Conference on Computer Vision and Pattern
  Recognition (CVPR)}, pages 224--231, June 2009.

\bibitem{hasler2009statistical}
Nils Hasler, Carsten Stoll, Martin Sunkel, Bodo Rosenhahn, and Hans-Peter
  Seidel.
\newblock A statistical model of human pose and body shape.
\newblock {\em Computer Graphics Forum}, 28(2):337--346, 2009.

\bibitem{hasson19_obman}
Yana Hasson, G{\"u}l Varol, Dimitrios Tzionas, Igor Kalevatykh, Michael~J.
  Black, Ivan Laptev, and Cordelia Schmid.
\newblock Learning joint reconstruction of hands and manipulated objects.
\newblock In {\em The IEEE Conference on Computer Vision and Pattern
  Recognition (CVPR)}, 2019.

\bibitem{stn2015}
Max Jaderberg, Karen Simonyan, Andrew Zisserman, and Koray Kavukcuoglu.
\newblock Spatial transformer networks.
\newblock In C. Cortes, N.~D. Lawrence, D.~D. Lee, M. Sugiyama, and R. Garnett,
  editors, {\em Advances in Neural Information Processing Systems}. 2015.

\bibitem{jiang2013hallucinated}
Yun Jiang, Hema Koppula, and Ashutosh Saxena.
\newblock Hallucinated humans as the hidden context for labeling 3d scenes.
\newblock In {\em The IEEE Conference on Computer Vision and Pattern
  Recognition (CVPR)}, pages 2993--3000, 2013.

\bibitem{Jiang:2012:objectArrangements}
Yun Jiang, Marcus Lim, and Ashutosh Saxena.
\newblock Learning object arrangements in 3d scenes using human context.
\newblock In {\em Proceedings of the 29th International Coference on
  International Conference on Machine Learning}, pages 907--914, 2012.

\bibitem{joo2018total}
Hanbyul Joo, Tomas Simon, and Yaser Sheikh.
\newblock Total capture: A 3{D} deformation model for tracking faces, hands,
  and bodies.
\newblock In {\em The IEEE Conference on Computer Vision and Pattern
  Recognition (CVPR)}, 2018.

\bibitem{kanazawa2017end}
Angjoo Kanazawa, Michael~J. Black, David~W. Jacobs, and Jitendra Malik.
\newblock End-to-end recovery of human shape and pose.
\newblock In {\em The IEEE Conference on Computer Vision and Pattern
  Recognition (CVPR)}, 2018.

\bibitem{shape2pose2014}
Vladimir~G Kim, Siddhartha Chaudhuri, Leonidas Guibas, and Thomas Funkhouser.
\newblock Shape2pose: Human-centric shape analysis.
\newblock {\em ACM Transactions on Graphics (TOG)}, 33(4):120, 2014.

\bibitem{Black_TrackPople}
Hedvig Kjellstr{\"o}m, Danica Kragi{\'c}, and Michael~J Black.
\newblock Tracking people interacting with objects.
\newblock In {\em The IEEE Conference on Computer Vision and Pattern
  Recognition (CVPR)}, pages 747--754, 2010.

\bibitem{koppula2013learning}
Hema~Swetha Koppula, Rudhir Gupta, and Ashutosh Saxena.
\newblock Learning human activities and object affordances from rgb-d videos.
\newblock {\em The International Journal of Robotics Research}, 32(8):951--970,
  2013.

\bibitem{kuo2001simple}
Arthur~D Kuo.
\newblock A simple model of bipedal walking predicts the preferred speed--step
  length relationship.
\newblock {\em Journal of biomechanical engineering}, 123(3):264--269, 2001.

\bibitem{kyriazis2013}
Nikolaos Kyriazis and Antonis Argyros.
\newblock Physically plausible {3D} scene tracking: The single actor
  hypothesis.
\newblock In {\em The IEEE Conference on Computer Vision and Pattern
  Recognition (CVPR)}, pages 9--16, 2013.

\bibitem{kyriazis2014}
Nikolaos Kyriazis and Antonis Argyros.
\newblock Scalable {3D} tracking of multiple interacting objects.
\newblock In {\em The IEEE Conference on Computer Vision and Pattern
  Recognition (CVPR)}, pages 3430--3437, 2014.

\bibitem{li2019motionforcesfromvideo}
Zongmian Li, Jiri Sedlar, Justin Carpentier, Ivan Laptev, Nicolas Mansard, and
  Josef Sivic.
\newblock Estimating 3d motion and forces of person-object interactions from
  monocular video.
\newblock In {\em The IEEE Conference on Computer Vision and Pattern
  Recognition (CVPR)}, 2019.

\bibitem{SMPL:2015}
Matthew Loper, Naureen Mahmood, Javier Romero, Gerard Pons-Moll, and Michael~J.
  Black.
\newblock {SMPL}: A skinned multi-person linear model.
\newblock {\em ACM Transactions on Graphics (TOG), (Proc. SIGGRAPH Asia)},
  34(6):248:1--248:16, Oct. 2015.

\bibitem{AMASS:2019}
Naureen Mahmood, Nima Ghorbani, Nikolaus F.~Troje, Gerard Pons-Moll, and
  Michael~J. Black.
\newblock {AMASS}: Archive of motion capture as surface shapes.
\newblock In {\em The IEEE International Conference on Computer Vision (ICCV)},
  Oct 2019.

\bibitem{vnect}
Dushyant Mehta, Srinath Sridhar, Oleksandr Sotnychenko, Helge Rhodin, Mohammad
  Shafiei, Hans-Peter Seidel, Weipeng Xu, Dan Casas, and Christian Theobalt.
\newblock Vnect: Real-time 3d human pose estimation with a single rgb camera.
\newblock {\em ACM Transactions on Graphics (TOG)}, 36(4):44:1--44:14, July
  2017.

\bibitem{Review_Moeslund_2006}
Thomas~B. Moeslund, Adrian Hilton, and Volker Kr\"{u}ger.
\newblock A survey of advances in vision-based human motion capture and
  analysis.
\newblock {\em Computer Vision and Image Understanding (CVIU)}, 104(2):90--126,
  2006.

\bibitem{iMapper2018}
Aron Monszpart, Paul Guerrero, Duygu Ceylan, Ersin Yumer, and Niloy~J Mitra.
\newblock imapper: interaction-guided scene mapping from monocular videos.
\newblock {\em ACM Transactions on Graphics (TOG)}, 38(4):92, 2019.

\bibitem{naseer2018indoorSurvey}
Muzammal Naseer, Salman Khan, and Fatih Porikli.
\newblock Indoor scene understanding in 2.5/3d for autonomous agents: A survey.
\newblock {\em IEEE Access}, 7:1859--1887, 2019.

\bibitem{nocedal2006nonlinear}
Jorge Nocedal and Stephen~J Wright.
\newblock {\em Nonlinear Equations}.
\newblock Springer, 2006.

\bibitem{Oikonomidis_1hand_object}
Iason Oikonomidis, Nikolaos Kyriazis, and Antonis~A. Argyros.
\newblock Full dof tracking of a hand interacting with an object by modeling
  occlusions and physical constraints.
\newblock In {\em The IEEE International Conference on Computer Vision (ICCV)},
  pages 2088--2095, 2011.

\bibitem{NBF:3DV:2018}
Mohamed Omran, Christoph Lassner, Gerard Pons-Moll, Peter~V. Gehler, and Bernt
  Schiele.
\newblock Neural body fitting: Unifying deep learning and model-based human
  pose and shape estimation.
\newblock In {\em 3DV}, Sept. 2018.

\bibitem{smplifyPP}
Georgios Pavlakos, Vasileios Choutas, Nima Ghorbani, Timo Bolkart, Ahmed A.~A.
  Osman, Dimitrios Tzionas, and Michael~J. Black.
\newblock Expressive body capture: 3d hands, face, and body from a single
  image.
\newblock In {\em The IEEE Conference on Computer Vision and Pattern
  Recognition (CVPR)}, 2019.

\bibitem{Pavlakos18}
Georgios Pavlakos, Luyang Zhu, Xiaowei Zhou, and Kostas Daniilidis.
\newblock Learning to estimate {3D} human pose and shape from a single color
  image.
\newblock In {\em The IEEE Conference on Computer Vision and Pattern
  Recognition (CVPR)}, 2018.

\bibitem{pham2018pami}
Tu-Hoa Pham, Nikolaos Kyriazis, Antonis~A Argyros, and Abderrahmane Kheddar.
\newblock Hand-object contact force estimation from markerless visual tracking.
\newblock {\em IEEE Transactions on Pattern Analysis and Machine Intelligence
  (TPAMI)}, 40(12):2883--2896, Dec 2018.

\bibitem{pirsiavash2012detecting}
Hamed Pirsiavash and Deva Ramanan.
\newblock Detecting activities of daily living in first-person camera views.
\newblock In {\em IEEE Conference on Computer Vision and Pattern Recognition
  (CVPR)}, pages 2847--2854, 2012.

\bibitem{Review_PoppeMotionAnalysis}
Ronald Poppe.
\newblock Vision-based human motion analysis: An overview.
\newblock {\em Computer Vision and Image Understanding (CVIU)}, 108(1-2):4--18,
  2007.

\bibitem{qi2017pointnet}
Charles~R Qi, Hao Su, Kaichun Mo, and Leonidas~J Guibas.
\newblock Pointnet: Deep learning on point sets for 3d classification and
  segmentation.
\newblock In {\em The IEEE Conference on Computer Vision and Pattern
  Recognition (CVPR)}, pages 652--660, 2017.

\bibitem{CAESAR}
Kathleen~M. Robinette, Sherri Blackwell, Hein Daanen, Mark Boehmer, Scott
  Fleming, Tina Brill, David Hoeferlin, and Dennis Burnsides.
\newblock {Civilian American and European Surface Anthropometry Resource
  (CAESAR)} final report.
\newblock Technical Report AFRL-HE-WP-TR-2002-0169, {US Air Force Research
  Laboratory}, 2002.

\bibitem{Rogez:ICCV:2015}
Gr{\'e}gory Rogez, James~S. Supan{\v{c}}i{\v{c}}~III, and Deva Ramanan.
\newblock Understanding everyday hands in action from rgb-d images.
\newblock In {\em The IEEE International Conference on Computer Vision (ICCV)},
  pages 3889--3897, 2015.

\bibitem{romero2017embodied}
Javier Romero, Dimitrios Tzionas, and Michael~J Black.
\newblock Embodied hands: Modeling and capturing hands and bodies together.
\newblock {\em ACM Transactions on Graphics (TOG)}, 36(6):245, 2017.

\bibitem{rosenhahn2008}
Bodo Rosenhahn, Christian Schmaltz, Thomas Brox, Joachim Weickert, Daniel
  Cremers, and Hans-Peter Seidel.
\newblock Markerless motion capture of man-machine interaction.
\newblock In {\em The IEEE Conference on Computer Vision and Pattern
  Recognition (CVPR)}, pages 1--8, June 2008.

\bibitem{ruenz2017cofusion}
Martin R{\"u}nz and Lourdes Agapito.
\newblock Co-fusion: Real-time segmentation, tracking and fusion of multiple
  objects.
\newblock In {\em 2017 IEEE International Conference on Robotics and Automation
  (ICRA)}, pages 4471--4478, 2017.

\bibitem{Sarafianos:Survey:2016}
Nikolaos Sarafianos, Bogdan Boteanu, Bogdan Ionescu, and Ioannis~A. Kakadiaris.
\newblock 3d human pose estimation: A review of the literature and analysis of
  covariates.
\newblock {\em Computer Vision and Image Understanding (CVIU)}, 152:1--20,
  2016.

\bibitem{scenegrok2014savva}
Manolis Savva, Angel~X Chang, Pat Hanrahan, Matthew Fisher, and Matthias
  Nie{\ss}ner.
\newblock Scenegrok: Inferring action maps in 3d environments.
\newblock {\em ACM Transactions on graphics (TOG)}, 33(6):212, 2014.

\bibitem{savva2016pigraphs}
Manolis Savva, Angel~X Chang, Pat Hanrahan, Matthew Fisher, and Matthias
  Nie{\ss}ner.
\newblock Pigraphs: learning interaction snapshots from observations.
\newblock {\em ACM Transactions on Graphics (TOG)}, 35(4):139, 2016.

\bibitem{simon2017hand}
Tomas Simon, Hanbyul Joo, Iain Matthews, and Yaser Sheikh.
\newblock Hand keypoint detection in single images using multiview
  bootstrapping.
\newblock In {\em The IEEE Conference on Computer Vision and Pattern
  Recognition (CVPR)}, 2017.

\bibitem{sridhar2016objectDexter}
Srinath Sridhar, Franziska Mueller, Michael Zollh{\"o}fer, Dan Casas, Antti
  Oulasvirta, and Christian Theobalt.
\newblock Real-time joint tracking of a hand manipulating an object from rgb-d
  input.
\newblock In {\em The European Conference on Computer Vision (ECCV)}, pages
  294--310, 2016.

\bibitem{Taylor:2017:ADF:3130800.3130853}
Jonathan Taylor, Vladimir Tankovich, Danhang Tang, Cem Keskin, David Kim,
  Philip Davidson, Adarsh Kowdle, and Shahram Izadi.
\newblock Articulated distance fields for ultra-fast tracking of hands
  interacting.
\newblock {\em ACM Transactions on Graphics (TOG)}, 36(6):244:1--244:12, Nov.
  2017.

\bibitem{collisionDeformableObjects}
Matthias Teschner, Stefan Kimmerle, Bruno Heidelberger, Gabriel Zachmann, Laks
  Raghupathi, Arnulph Fuhrmann, Marie-Paule Cani, Fran{\c{c}}ois Faure, Nadia
  Magnenat-Thalmann, Wolfgang Strasser, and Pascal Volino.
\newblock Collision detection for deformable objects.
\newblock In {\em Eurographics}, pages 119--139, 2004.

\bibitem{Tsoli:2018:ECCV}
Aggeliki Tsoli and Antonis~A. Argyros.
\newblock Joint 3d tracking of a deformable object in interaction with a hand.
\newblock In {\em The European Conference on Computer Vision (ECCV)}, 2018.

\bibitem{Tzionas:IJCV:2016}
Dimitrios Tzionas, Luca Ballan, Abhilash Srikantha, Pablo Aponte, Marc
  Pollefeys, and Juergen Gall.
\newblock Capturing hands in action using discriminative salient points and
  physics simulation.
\newblock {\em International Journal of Computer Vision (IJCV)},
  118(2):172--193, 2016.

\bibitem{Vondrak2013}
Marek Vondrak, Leonid Sigal, and Odest~Chadwicke Jenkins.
\newblock Dynamical simulation priors for human motion tracking.
\newblock {\em IEEE Transactions on Pattern Analysis and Machine Intelligence
  (TPAMI)}, 35(1):52--65, Jan 2013.

\bibitem{wei2016convolutional}
Shih-En Wei, Varun Ramakrishna, Takeo Kanade, and Yaser Sheikh.
\newblock Convolutional pose machines.
\newblock In {\em The IEEE Conference on Computer Vision and Pattern
  Recognition (CVPR)}, 2016.

\bibitem{Yamamoto2000}
Masanobu Yamamoto and Katsutoshi Yagishita.
\newblock Scene constraints-aided tracking of human body.
\newblock In {\em The IEEE Conference on Computer Vision and Pattern
  Recognition (CVPR)}, volume~1, pages 151--156 vol.1, June 2000.

\bibitem{yao2010modeling}
Bangpeng Yao and Li Fei-Fei.
\newblock Modeling mutual context of object and human pose in human-object
  interaction activities.
\newblock In {\em IEEE Conference on Computer Vision and Pattern Recognition
  (CVPR)}, pages 17--24, 2010.

\bibitem{zanfir2018monocular}
Andrei Zanfir, Elisabeta Marinoiu, and Cristian Sminchisescu.
\newblock Monocular 3d pose and shape estimation of multiple people in natural
  scenes-the importance of multiple scene constraints.
\newblock In {\em Proceedings of the IEEE Conference on Computer Vision and
  Pattern Recognition (CVPR)}, pages 2148--2157, 2018.

\bibitem{zhao2004}
Tao Zhao and Ram Nevatia.
\newblock Tracking multiple humans in complex situations.
\newblock {\em IEEE Transactions on Pattern Analysis and Machine Intelligence
  (TPAMI)}, 26(9):1208--1221, Sep. 2004.

\bibitem{open3D}
Qian-Yi Zhou, Jaesik Park, and Vladlen Koltun.
\newblock {Open3D}: {A} modern library for {3D} data processing.
\newblock {\em arXiv:1801.09847}, 2018.

\bibitem{zollhofer2018sotaRconstructionRGBD}
Michael Zollh{\"o}fer, Patrick Stotko, Andreas G{\"o}rlitz, Christian Theobalt,
  Matthias Nie{\ss}ner, Reinhard Klein, and Andreas Kolb.
\newblock State of the art on 3d reconstruction with rgb-d cameras.
\newblock {\em Computer Graphics Forum}, 37(2):625--652, 2018.

\end{thebibliography}


\begin{thebibliography}{1}\itemsep=-1pt

\bibitem{Eigen:2014}
David Eigen, Christian Puhrsch, and Rob Fergus.
\newblock Depth map prediction from a single image using a multi-scale deep
  network.
\newblock In {\em Advances in Neural Information Processing Systems}, pages
  2366–--2374, 2014.

\bibitem{naseer2018indoorSurvey}
Muzammal Naseer, Salman Khan, and Fatih Porikli.
\newblock Indoor scene understanding in 2.5/3d for autonomous agents: A survey.
\newblock {\em IEEE Access}, 7:1859--1887, 2019.

\bibitem{smplifyPP}
Georgios Pavlakos, Vasileios Choutas, Nima Ghorbani, Timo Bolkart, Ahmed A.~A.
  Osman, Dimitrios Tzionas, and Michael~J. Black.
\newblock Expressive body capture: 3d hands, face, and body from a single
  image.
\newblock In {\em The IEEE Conference on Computer Vision and Pattern
  Recognition (CVPR)}, 2019.

\bibitem{savva2016pigraphs}
Manolis Savva, Angel~X Chang, Pat Hanrahan, Matthew Fisher, and Matthias
  Nie{\ss}ner.
\newblock Pigraphs: learning interaction snapshots from observations.
\newblock {\em ACM Transactions on Graphics (TOG)}, 35(4):139, 2016.

\end{thebibliography}
}

\clearpage
\includepdf[pages=1]{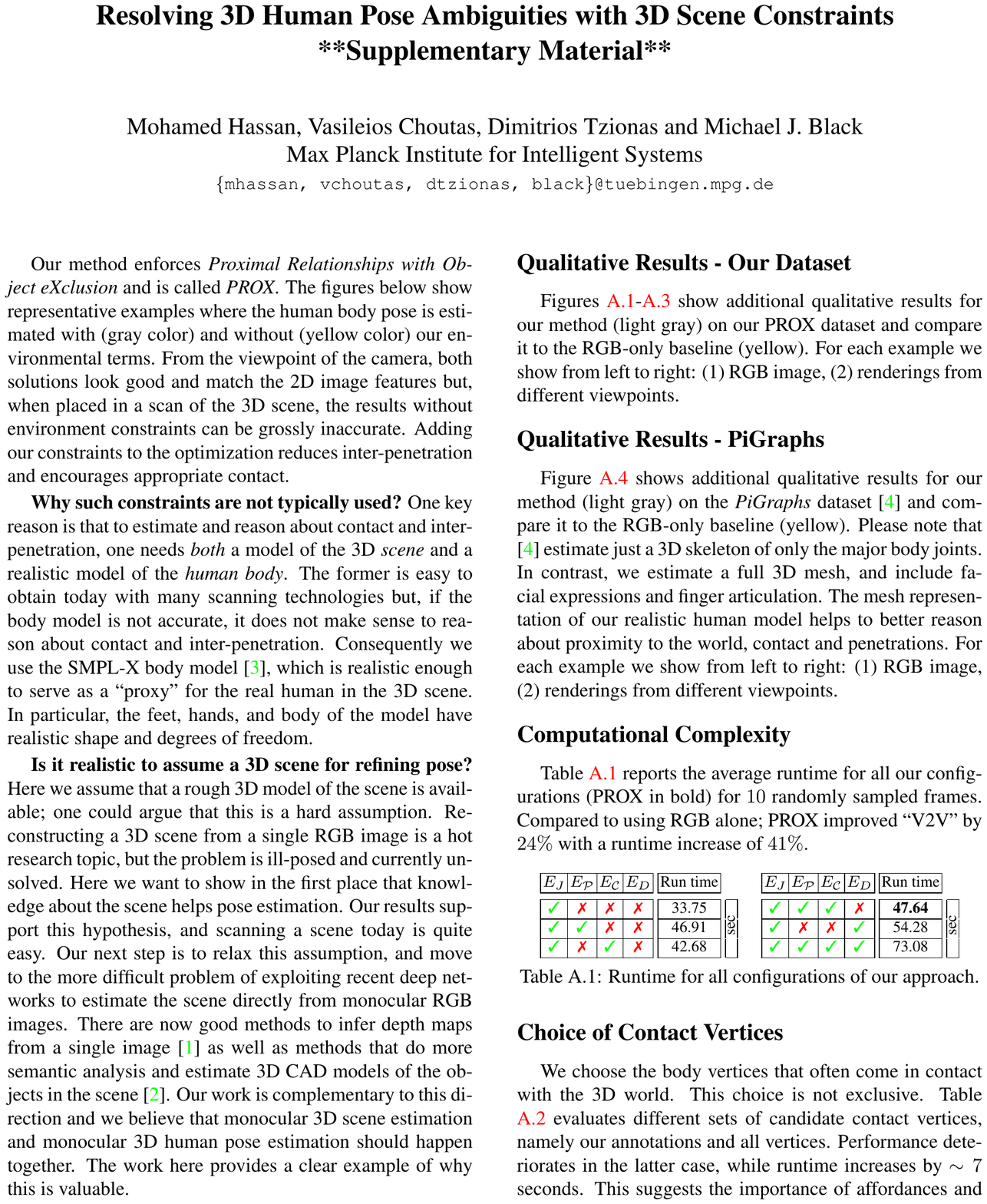}
\includepdf[pages=2]{./iccv2019_human_world___SUPPLEMENTARY___00__COMPRESSED.pdf}
\includepdf[pages=3]{./iccv2019_human_world___SUPPLEMENTARY___00__COMPRESSED.pdf}
\includepdf[pages=4]{./iccv2019_human_world___SUPPLEMENTARY___00__COMPRESSED.pdf}
\includepdf[pages=5]{./iccv2019_human_world___SUPPLEMENTARY___00__COMPRESSED.pdf}
\includepdf[pages=6]{./iccv2019_human_world___SUPPLEMENTARY___00__COMPRESSED.pdf}
\includepdf[pages=7]{./iccv2019_human_world___SUPPLEMENTARY___00__COMPRESSED.pdf}
\end{document}


\title{Resolving 3D Human Pose Ambiguities with 3D Scene Constraints\\***Supplementary Material**}

\author{Mohamed Hassan, Vasileios Choutas, Dimitrios Tzionas and Michael J. Black\\
	Max Planck Institute for Intelligent Systems\\
	{\tt\small \{mhassan, vchoutas, dtzionas, black\}@tuebingen.mpg.de}
}
\maketitle
\ificcvfinal\thispagestyle{empty}\fi

Our method enforces {\em Proximal Relationships with Object eXclusion} and is called {\em \prox}.
The figures below show representative examples where the human body pose is estimated with (gray color) and without (yellow color) our environmental terms.
From the viewpoint of the camera, both solutions look good and match the \twoD image features but, when placed in a scan of the \threeD scene, the results without environment constraints can be grossly inaccurate.
Adding our constraints to the optimization reduces inter-penetration and encourages appropriate contact.

{\bf Why such constraints are not typically used?}
One key reason is that to estimate and reason about contact and inter-penetration, one needs \emph{both} a model of the \threeD \emph{scene} and a realistic model of the \emph{human body}.
The former is easy to obtain today with many scanning technologies but, if the body model is not accurate, it does not make sense to reason about contact and inter-penetration.
Consequently we use the \smplHF body model \cite{smplifyPP}, which is realistic enough to serve as a ``proxy'' for the real human in the \threeD scene.
In particular, the feet, hands, and body of the model have realistic shape and degrees of freedom.

{\bf Is it realistic to assume a \threeD scene for refining pose?}
Here we assume that a rough \threeD model of the scene is available; one could argue that this is a hard assumption.
Reconstructing a \threeD scene from a single RGB image is a hot research topic, but the problem is ill-posed and currently unsolved.
Here we want to show in the first place that knowledge about the scene helps pose estimation.
Our results support this hypothesis, and scanning a scene today is quite easy.
Our next step is to relax this assumption, and move to the more difficult problem of exploiting recent deep networks to estimate the scene directly from monocular \rgb images.
There are now good methods to infer depth maps from a single image \cite{Eigen:2014} as well as methods that do more semantic analysis and estimate \threeD CAD models of the objects in the scene \cite{naseer2018indoorSurvey}.
Our work is complementary to this direction and we believe that monocular \threeD scene estimation and monocular \threeD human pose estimation should happen together.
The work here provides a clear example of why this is valuable.

\newcommand{\captionQualitativeOurs}{
											Qualitative results on our \prox dataset.
											The human body pose is estimated \emph{with} (light gray) and \emph{without} (yellow) our environmental terms.
											We show from left to right:
											(1) RGB images,
											(2) renderings from different viewpoints.
}
\newcommand{\captionQualitativePigraph}{
											Qualitative results on the \pigraph \cite{savva2016pigraphs} dataset.
											The human body pose is estimated \emph{with} (gray color) and \emph{without} (yellow color) our environmental terms.
											Please note that \cite{savva2016pigraphs} estimate just a \threeD skeleton of only the major body joints.
											We show from left to right:
											(1) RGB images,
											(2) renderings from different viewpoints.
}
\newcommand{\captionFailureCases}{
											Representative failure cases on our \prox dataset.
											We show from left to right:
											(1) RGB image,
											(2) \openpose result overlayed on the RGB image,
											(3) result of our method.
}

\newcommand{\sizQualitativeOurs}{0.98}
\newcommand{\sizQualitativePigraph}{0.98}
\newcommand{\sizFailureCases}{0.98}

\section*{Qualitative Results - Our Dataset}
Figures \ref{fig:supMat_qualitativeOURS_01}-\ref{fig:supMat_qualitativeOURS_03} show additional qualitative results for our method (light gray) on our \prox dataset and compare it to the \rgb-only baseline (yellow).
For each example we show from left to right:
(1) RGB image,
(2) renderings from different viewpoints.

\section*{Qualitative Results - \pigraph}
Figure \ref{fig:supMat_qualitativePIGRAPHS_01} shows additional qualitative results for our method (light gray) on the \emph{\pigraph} dataset \cite{savva2016pigraphs} and compare it to the \rgb-only baseline (yellow).
Please note that \cite{savva2016pigraphs} estimate just a \threeD skeleton of only the major body joints.
In contrast, we estimate a full \threeD mesh, and include facial expressions and finger articulation.
The mesh representation of our realistic human model helps to better reason about proximity to the world, contact and penetrations.
For each example we show from left to right:
(1) RGB image,
(2) renderings from different viewpoints.

\section*{Computational Complexity}
Table \ref{tab:computational_complexity} reports the average runtime for all our configurations (\prox in bold) for $10$ randomly sampled frames. Compared to using \rgb alone; \prox improved ``V2V'' by $24\%$ with a runtime increase of $41\%$.
\begin{table}[h!]
	\begin{center}
	\footnotesize
	\setlength{\tabcolsep}{1pt}
	\begin{tabular}{|c|c|c|c|c|c|c|c|}																																								\hhline{----~-~~}
	$E_J$	& $E_\contact$	&	$E_{\mColl}$	&	$E_D$	&	\multicolumn{1}{c}{}		&	\multicolumn{1}{|c|}{Run time}	&	\multicolumn{1}{c}{}	&	\multicolumn{1}{c}{}							\\	\hhline{----~-~~}
	\noalign{\smallskip}																																												\hhline{----~-~-}
	\cmark	&  \xmark		&	\xmark		&	\xmark	&	\multicolumn{1}{c}{}		&	\multicolumn{1}{|c|}{33.75}		&	{}	&	\multirow{3}{*}{\centering\begin{turn}{90}sec\end{turn}}		\\	\hhline{----~-~~}
	\cmark	&  \cmark		&	\xmark		&	\xmark	&	\multicolumn{1}{c}{}		&	\multicolumn{1}{|c|}{46.91}		&	{}	&	{}															\\	\hhline{----~-~~}
	\cmark	&  \xmark		&	\cmark		&	\xmark	&	\multicolumn{1}{c}{}		&	\multicolumn{1}{|c|}{42.68}		&	{}	&	{}															\\	\hhline{----~-~-}
	\end{tabular}
	\hspace{1.0em}
	\begin{tabular}{|c|c|c|c|c|c|c|c|}																																								\hhline{----~-~~}
	$E_J$	& $E_\contact$	&	$E_{\mColl}$	&	$E_D$	&	\multicolumn{1}{c}{}		&	\multicolumn{1}{|c|}{Run time}	&	\multicolumn{1}{c}{}	&	\multicolumn{1}{c}{}							\\	\hhline{----~-~~}
	\noalign{\smallskip}																																												\hhline{----~-~-}
	\cmark	&  \cmark		&	\cmark		&	\xmark	&	\multicolumn{1}{c}{}		&	\multicolumn{1}{|c|}{{\bf 47.64}}&	{}	&	\multirow{3}{*}{\centering\begin{turn}{90}sec\end{turn}}		\\	\hhline{----~-~~}
	\cmark	&  \xmark		&	\xmark		&	\cmark	&	\multicolumn{1}{c}{}		&	\multicolumn{1}{|c|}{54.28}		&	{}	&	{}															\\	\hhline{----~-~~}
	\cmark	&  \cmark		&	\cmark		&	\cmark	&	\multicolumn{1}{c}{}		&	\multicolumn{1}{|c|}{73.08}		&	{}	&	{}															\\	\hhline{----~-~-}
	\end{tabular}
	\end{center}
	\vspace{-1.6em}
	\caption{
				Runtime for all configurations of our approach.
	}
	\label{tab:computational_complexity}
	\vspace{-0.8em}
\end{table}

\section*{Choice of Contact Vertices}
We choose the body vertices that often come in contact with the \threeD world.
This choice is not exclusive.
Table \ref{tab:all_verts_contact} evaluates different sets of candidate contact vertices,
namely
our annotations and all vertices.
Performance deteriorates in the latter case, while runtime increases by $\sim 7$ seconds.
This suggests the importance of affordances and semantics; %
future work can learn the likely contact vertices for different object classes in a data-driven fashion.
To this end, the community first needs training data similar to the data generated by our work.
\vspace{-0.4em}
\begin{table}[h!]
	\begin{center}
	\footnotesize
	\setlength{\tabcolsep}{1.0pt}
	\begin{tabular}{|c|c|c|c|c|c|c|c|}																																																											\hhline{-~----~~}
	\multicolumn{1}{|c|}{Contact vertices}	& \multicolumn{1}{c}{}	& \multicolumn{1}{|c|}{PJE}		& \multicolumn{1}{c|}{V2V}		&	\multicolumn{1}{c|}{p.PJE}	&	\multicolumn{1}{c|}{p.V2V}																			\\	\hhline{-~----~~}
	\noalign{\smallskip}																																																															\hhline{-~----~-}
	\multicolumn{1}{|c|}{Selected of Fig. 2}	& \multicolumn{1}{c}{}	& \multicolumn{1}{|c|}{$208.03$}	& \multicolumn{1}{c|}{$208.57$}	&	\multicolumn{1}{c|}{$72.76$}	&	\multicolumn{1}{c|}{$60.95$} 	&	{}	& \multirow{2}{*}{\centering\begin{turn}{90}mm\end{turn}}		\\	\hhline{-~----~~}
	\multicolumn{1}{|c|}{All selected}		& \multicolumn{1}{c}{}	& \multicolumn{1}{|c|}{$217.82$}	& \multicolumn{1}{c|}{$216.62$}	&	\multicolumn{1}{c|}{$72.35$}	&	\multicolumn{1}{c|}{$60.16$}		&	{}	&															\\	\hhline{-~----~-}
	\end{tabular}
	\end{center}
	\vspace{-1.6em}
	\caption{
				Different sets of candidate contact vertices.
	}
	\label{tab:all_verts_contact}
	\vspace{-0.8em}
\end{table}

\section*{Failure Cases}
Figures \ref{fig:supMat_qualitativeFAILURES_01}-\ref{fig:supMat_qualitativeFAILURES_02} show failure cases of our method (light gray) on our \prox dataset.
For each example we show from left to right:
(1) RGB image,
(2) \openpose result overlayed on the RGB image,
(3) result of our method.
Figure \ref{fig:supMat_qualitativeFAILURES_01}-top shows that our method still results in some penetration.
Our assumption of a static scene is not always true; in this case the bed is deformable and its shape changes during interaction.
In future work we plan to model deformations of the human body and the world.
Figure \ref{fig:supMat_qualitativeFAILURES_01}-bottom shows a failure of our inter-penetration term.
In cases where initialization of body translation is not accurate enough, the optimizer might end up in a local minimum that is not always in agreement with the real pose in \threeD space.
Figure \ref{fig:supMat_qualitativeFAILURES_02} shows typical failure cases of \openpose.
In Figure \ref{fig:supMat_qualitativeFAILURES_02}-top the left leg is not detected correctly, while
in Figure \ref{fig:supMat_qualitativeFAILURES_02}-middle and Figure \ref{fig:supMat_qualitativeFAILURES_02}-bottom
several body joints are flipped by \openpose.

\begin{figure*}
    \centering
    \subfloat{	\includegraphics[trim=000mm 000mm 000mm 000mm, clip=false, width=0.88 \linewidth]{./figures_SUP_MAT/qualitative_examples_15_.jpg}	}	\\	\vspace*{-0.5em}
    \subfloat{	\includegraphics[trim=000mm 000mm 000mm 000mm, clip=false, width=0.88 \linewidth]{./figures_SUP_MAT/qualitative_examples_02.jpg}	}	\\	\vspace*{-0.5em}
    \subfloat{	\includegraphics[trim=000mm 000mm 000mm 000mm, clip=false, width=0.88 \linewidth]{./figures_SUP_MAT/qualitative_examples_11.jpg}	}	\\	\vspace*{-0.5em}
    \subfloat{	\includegraphics[trim=000mm 000mm 000mm 000mm, clip=false, width=0.88 \linewidth]{./figures_SUP_MAT/qualitative_examples_12_.jpg}	}
    \caption{
        			\captionQualitativeOurs
    }
    \label{fig:supMat_qualitativeOURS_01}
\end{figure*}
\begin{figure*}
    \centering
    \subfloat{	\includegraphics[trim=000mm 000mm 000mm 000mm, clip=false, width=0.90 \linewidth]{./figures_SUP_MAT/qualitative_examples_09_.jpg}	}	\\	\vspace*{-0.5em}
    \subfloat{	\includegraphics[trim=000mm 000mm 000mm 000mm, clip=false, width=0.90 \linewidth]{./figures_SUP_MAT/qualitative_examples_14_.jpg}	}	\\	\vspace*{-0.5em}
    \subfloat{	\includegraphics[trim=000mm 000mm 000mm 000mm, clip=false, width=0.90 \linewidth]{./figures_SUP_MAT/qualitative_examples_13_.jpg}	}
    \caption{
        			\captionQualitativeOurs
    }
    \label{fig:supMat_qualitativeOURS_02}
\end{figure*}
\begin{figure*}
    \centering
    \subfloat{	\includegraphics[trim=000mm 000mm 000mm 000mm, clip=false, width=0.80 \linewidth]{./figures_SUP_MAT/qualitative_examples_10.jpg}	}	\\	\vspace*{-0.5em}
    \subfloat{	\includegraphics[trim=000mm 000mm 000mm 000mm, clip=false, width=\sizQualitativeOurs \linewidth]{./figures_SUP_MAT/qualitative_examples_04.jpg}	}
    \caption{
        			\captionQualitativeOurs
    }
    \label{fig:supMat_qualitativeOURS_03}
\end{figure*}

\begin{figure*}
    \centering
    \subfloat{	\includegraphics[trim=000mm 000mm 000mm 000mm, clip=false, width=0.88 \linewidth]{./figures_SUP_MAT/pigraphs_01.jpg}	}	\\	\vspace*{-0.5em}
    \subfloat{	\includegraphics[trim=000mm 000mm 000mm 000mm, clip=false, width=0.88 \linewidth]{./figures_SUP_MAT/pigraphs_02.jpg}	}
    \caption{
        			\captionQualitativePigraph
    }
    \label{fig:supMat_qualitativePIGRAPHS_01}
\end{figure*}

\begin{figure*}
    \centering
    \subfloat{	\includegraphics[trim=000mm 000mm 000mm 090mm, clip=true, width=0.88 \linewidth]{./figures_SUP_MAT/failure_cases_01.jpg}	}	\\	\vspace*{-0.5em}
    \subfloat{	\includegraphics[trim=000mm 000mm 000mm 020mm, clip=true, width=0.88 \linewidth]{./figures_SUP_MAT/failure_cases_02.jpg}	}
    \caption{
        			\captionFailureCases
    }
    \label{fig:supMat_qualitativeFAILURES_01}
\end{figure*}
\begin{figure*}
    \centering
    \subfloat{	\includegraphics[trim=000mm 000mm 000mm 000mm, clip=false, width=0.94 \linewidth]{./figures_SUP_MAT/failure_cases_03_.jpg}	}	\\	\vspace*{-0.5em}
    \subfloat{	\includegraphics[trim=000mm 000mm 000mm 000mm, clip=false, width=0.94 \linewidth]{./figures_SUP_MAT/failure_cases_04.jpg}	}	\\	\vspace*{-0.5em}
    \subfloat{	\includegraphics[trim=000mm 000mm 000mm 000mm, clip=false, width=0.94 \linewidth]{./figures_SUP_MAT/failure_cases_05.jpg}	}
    \caption{
        			\captionFailureCases
    }
    \label{fig:supMat_qualitativeFAILURES_02}
\end{figure*}

{\small
\bibliographystyle{ieee_fullname}
\bibliography{egbib,bib_bodies,bib_handrefs,bib_faces,bib_vposer,bib_homogenus}
}